%% file: main.tex
\documentclass[sigconf, screen]{acmart}
\AtBeginDocument{%
  }

\setcopyright{acmlicensed}
\copyrightyear{2026}
\acmYear{2026}
\acmDOI{XXXXXXX.XXXXXXX}
\acmConference[Conference acronym 'XX]{Make sure to enter the correct
  conference title from your rights confirmation email}{June 03--05,
  2018}{Woodstock, NY}
\acmISBN{978-1-4503-XXXX-X/2018/06}

\usepackage[utf8]{inputenc}

\usepackage{xcolor}
\usepackage{graphicx}
\usepackage{subcaption}
\usepackage{tabularx}
\usepackage{booktabs}
\usepackage{multirow}
\usepackage{arydshln}
\usepackage{enumitem}
\usepackage{tcolorbox}
\usepackage{courier}

\definecolor{promptBlue}{RGB}{30, 64, 175}
\definecolor{ruleGray}{RGB}{107, 114, 128}
\definecolor{bgGray}{RGB}{249, 250, 251}
\definecolor{obsGreen}{RGB}{20, 70, 40}
\definecolor{lightGreen}{RGB}{245, 250, 245}
\definecolor{alertRed}{RGB}{180, 0, 0}
\definecolor{planGray}{RGB}{230,230,230}
\definecolor{codeBg}{RGB}{245,245,245} 
\definecolor{logicBlue}{RGB}{40,90,160} 
\definecolor{warnRed}{RGB}{200,50,50}

\tcbuselibrary{skins, breakable}

\newtcolorbox{systemprompt}{
    enhanced,
    breakable,
    colback=white,
    colframe=promptBlue,
    coltitle=white,
    fonttitle=\bfseries\large,
    title=SYSTEM ROLE: MEMORY AGENT,
    attach boxed title to top left={yshift=-3mm, xshift=3mm},
    boxed title style={colback=promptBlue, sharp corners},
    boxrule=1pt,
    top=4mm,
    bottom=2mm
}
\tcbuselibrary{skins, breakable}

\newtcolorbox{promptbox}[1]{
    colback=white,
    colframe=brandblue,
    fonttitle=\bfseries\large,
    title=#1,
    sharp corners,
    boxrule=0.8pt,
    enhanced,
    attach boxed title to top left={yshift=-2mm, xshift=2mm},
    boxed title style={colback=brandblue},
    breakable
}
\begin{document}

\title{MGA: Memory-Driven GUI Agent for Observation-Centric Interaction}

\author{Weihua Cheng}
\affiliation{%
  \institution{Shanghai Tech University}
  \city{Shanghai}
  \country{China}
}
\email{chengwh2024@shanghaitech.edu.cn}

\author{Junming Liu}
\affiliation{%
  \institution{Tongji University}
  \city{Shanghai}
  \country{China}
}
\email{2331890@tongji.edu.cn}

\author{Yifei Sun}
\affiliation{%
 \institution{East China University of Science and Technology}
 \city{Shanghai}
\country{China}}
\email{y21220030@mail.ecust.edu.cn}

\author{Botian Shi}
\affiliation{%
 \institution{Shanghai AI Laboratory}
 \city{Shanghai}
\country{China}}
\email{shibotian@pjlab.org.cn}

\author{Yirong Chen}
\authornote{Corresponding author.}
\affiliation{%
 \institution{Shanghai AI Laboratory}
 \city{Shanghai}
\country{China}}
\email{chenyirong@pjlab.org.cn}

\author{Ding Wang}
\authornote{Corresponding author.}
\affiliation{%
 \institution{Shanghai AI Laboratory}
 \city{Shanghai}
\country{China}}
\email{wangding@pjlab.org.cn}

\renewcommand{\shortauthors}{Cheng et al.}

\begin{abstract}
Multimodal Large Language Models (MLLMs) have significantly advanced GUI agents, yet long-horizon automation remains constrained by two critical bottlenecks: context overload from raw sequential trajectory dependence and architectural redundancy from over-engineered expert modules. Prevailing End-to-End and Multi-Agent paradigms struggle with error cascades caused by concatenated visual-textual histories and incur high inference latency due to redundant expert components, limiting their practical deployment. To address these issues, we propose the Memory-Driven GUI Agent (MGA), a minimalist framework that decouples long-horizon trajectories into independent decision steps linked by a structured state memory. MGA operates on an ``Observe First and Memory Enhancement'' principle, powered by two tightly coupled core mechanisms: (1) an Observer module that acts as a task-agnostic, intent-free screen state reader to eliminate confirmation bias, visual hallucinations, and perception bias at the root; and (2) a Structured Memory mechanism that distills, validates, and compresses each interaction step into verified state deltas, constructing a lightweight state transition chain to avoid irrelevant historical interference and system redundancy. By replacing raw historical aggregation with compact, fact-based memory transitions, MGA drastically reduces cognitive overhead and system complexity. Extensive experiments on OSWorld and real-world applications demonstrate that MGA achieves highly competitive performance in open-ended GUI tasks while maintaining architectural simplicity, offering a scalable and efficient blueprint for next-generation GUI automation. \textcolor{blue}{https://github.com/MintyCo0kie/MGA4OSWorld}.
\end{abstract}

\ccsdesc[500]{Computing methodologies~Intelligent agents}
\ccsdesc[400]{Human-centered computing~Graphical user interfaces}
\ccsdesc[300]{Computing methodologies~Natural language processing}
\ccsdesc[300]{Computing methodologies~Computer vision}
\ccsdesc[200]{Computing methodologies~Planning and scheduling}

\keywords{GUI automation, multimodal large language models, structured memory, long-horizon planning, observer-first perception, active error correction}

\received{20 February 2007}
\received[revised]{12 March 2009}
\received[accepted]{5 June 2009}

\maketitle

\section{Introduction}
\label{sec:intro}
\input{tex/intro}

\section{Related Work}
\label{sec:related work}
\input{tex/related}

\section{Methodology}
\label{sec:methodology}
\input{tex/method}

\section{Experiment}
\label{sec:exp}

\input{tex/exp}

\section{Conclusion}
\label{sec:conclusion}
\input{tex/conclution}

\bibliographystyle{splncs04}
\bibliography{main}

\input{appendix}
\end{document}

%% file: tex/intro.tex
Driven by the rapid advancement of Multimodal Large Language Models (MLLMs) and agent frameworks, Graphical User Interface (GUI) agents have evolved from rule-driven approaches to complex systems capable of autonomous, human-like perception and interaction~\cite{yin2024survey,wang2024gui,agashe2024agent,caffagni2024revolution,zhai2024fine,bai2025qwen2}. However, long-horizon GUI tasks remain highly challenging due to the partial observability and dynamic layouts inherent to the GUI environment ~\cite{nguyen2025gui,agashe2024agent,zhang2024large}. Existing approaches primarily fall into two paradigms: End-to-End (E2E) models and Modular Multi-Agent Systems (MAS). E2E models, such as UI-TARS~\cite{qin2025ui} and Claude-Sonnet~\cite{anthropic20253}, directly map raw screenshots to pixel-level actions, but struggle with accumulating historical context. In contrast, MAS frameworks decompose functionality into specialized agents for planning, grounding, and execution~\cite{wang2024gui,nguyen2024gui,cheng2024seeclick,song2025coact}, improving modularity at the cost of increased coordination overhead.

\begin{figure*}[h]
  \centering
\includegraphics[width=0.8\textwidth]{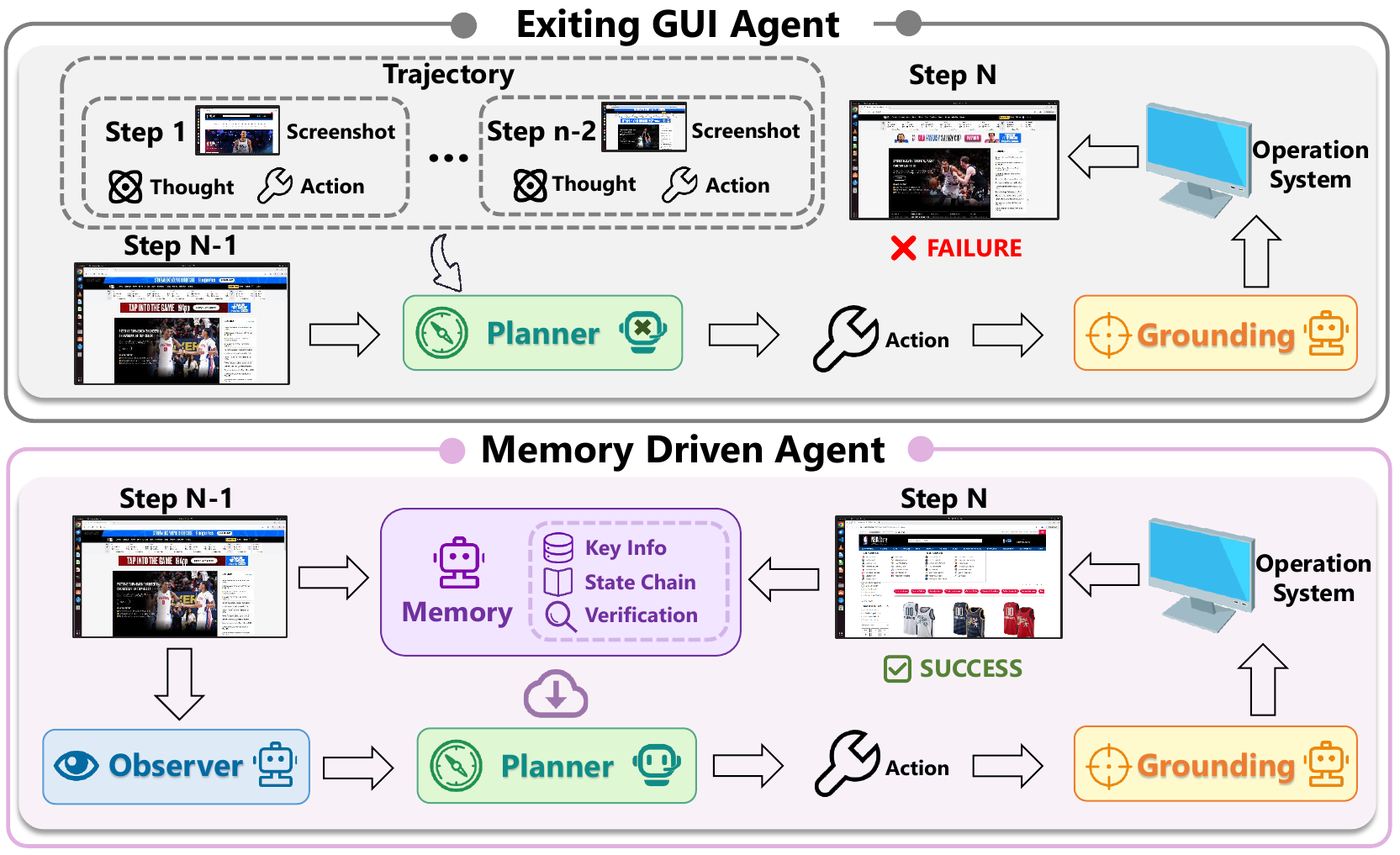}
  \caption{Comparative illustration of GUI agent workflows. Top: Conventional paradigm aggregates historical screenshots, decision chains, and action trajectories into a single context, leading to attention dilution and error accumulation. Bottom: MGA decouples redundant trajectories via structured memory, enabling a streamlined pipeline that preserves focused attention with observation-first for long-horizon tasks.}
  \label{fig:gui-agent-flow}
\end{figure*}

Despite these advances, existing paradigms for long-horizon GUI interaction~\cite{xie2024osworld,agashe2025agent,jiang2025screencoder} still face two fundamental bottlenecks that limit their practical deployment: 1) Mainstream approaches model GUI task execution purely as flat sequential trajectories~\cite{shaw2023pixels,li2023zero}, concatenating historical screenshots, plans, and actions into the context and requiring the model to rely on the entire raw trajectory to infer the current state (Fig.~\ref{fig:gui-agent-flow}). This design leaves the model susceptible to misguidance from irrelevant or erroneous history, impairing its ability to track the current state. 2) To operate in open-ended environments, state-of-the-art frameworks increasingly rely on large-scale, fully-featured expert agents. For example, OS-Symphony and Agent S3~\citep{yang2026ossymphonyholisticframeworkrobust,gonzalez2025unreasonable} integrate multiple MLLMs with orchestrators, reflection modules, and tool chains. While seemingly effective, such designs introduce redundancy and complexity, constituting over-design for the majority of routine GUI tasks.

To address the above two core bottlenecks, we revisit GUI automation from first principles and advocate reconceptualizing the core reasoning unit of GUI agents. We argue that the fusion of observation and memory forms a sufficient representation of the latent GUI state. We propose the Memory-Driven GUI Agent (MGA), which provides a novel solution to the aforementioned limitations: historical interference from raw trajectory dependence and redundancy caused by over-complex system design. MGA is built upon two tightly coupled core modules: \textit{\textbf{Observer}} and \textit{\textbf{Memory}}. 

Specifically,to eliminate perception bias at its root, we introduce the Observer module as a task-agnostic, intent-free screen state reader. Existing GUI agents typically perform perception under the guidance of task intent, leading to confirmation bias where models over-focus on actionable UI elements while neglecting critical contextual information such as textual descriptions and background states~\cite{dong2025say, wanyan2025look}. This coupling between perception and decision-making produces incomplete and biased state representations, which prior work attempts to compensate for by adding complex expert modules. In contrast, the Observer in MGA is strictly prohibited from any planning or intent inference.
It operates exclusively in a task-agnostic perception mode and performs only objective and fine-grained state characterization. This constraint, which enforces clarification of current factual states before accessing historical and task information, fundamentally eliminates visual hallucinations caused by subjective model inference and preserves crucial background semantics. It thus provides a pure and unbiased initial state benchmark for downstream decision-making modules, as well as a reliable input source for the paired memory mechanism.

Besides, the Memory module operates in tight coordination with the Observer to achieve sufficient GUI state representation. Rather than retaining unprocessed raw interaction trajectories, it distills, validates, and compresses each step’s action and execution effect into verified state increments to construct a lightweight, append-only state transition chain decoupling long-horizon interaction trajectories into independent decision steps connected via state memory. By eliminating redundant historical burdens and interference from irrelevant or erroneous history, memory enabling reliable backtracking for the planner also retains essential contextual information including task constraints, critical verified results, and loop-monitoring anomaly signals throughout execution to avoid system redundancy and complexity and mitigate perception bias at its source, thus ensuring the memory remains compact, dependable, and efficiently accessible for real time decision-making.

In summary, this work makes the following core contributions:
\begin{itemize}
    \item We reframe long-horizon GUI automation as independent yet logically connected decision nodes, effectively alleviating context overload caused by raw historical sequences.
    \item We propose MGA, an "observation first + memory enhancement" GUI agent. It leverages a task-agnostic observer to reduce visual hallucinations, and a Structured Memory mechanism to prevent behavioral stagnation and error cascades.
    \item Experiments on OSWorld demonstrate that MGA maintains highly competitive performance in open-ended GUI tasks .
\end{itemize}

%% file: tex/related.tex
\subsection{End-to-End and Specialist GUI Models} 

The end-to-end paradigm integrates perception, reasoning, and execution into a unified architecture, directly mapping screen observations and user instructions to executable actions. This paradigm emphasizes model coherence, training efficiency, and cross-domain generalization~\citep{hong2024cogagent,ye2025mobile,xu2024aguvis,liu2025infiguiagent}. Recent studies further explore temporal dynamics and multimodal interaction for GUI agents.
ScaleTrack~\citep{huang2025scaletrack} reconstructs historical trajectories and predicts future actions to capture the temporal evolution of GUI state-action dependencies. UITron-Speech~\citep{han2025guirobotron} introduces the first speech-driven GUI agent, addressing challenges in multimodal instruction understanding and visual grounding.
Representative domain-specialist models include the UI-TARS series~\citep{qin2025ui}, which pursues unified action modeling across applications. UI-TARS incorporates System-2 reasoning and explicit deliberation, while its follow-up UI-TARS-2 further integrates multi-round reinforcement learning and file system access for hybrid desktop/OS environments. The OpenCUA series~\citep{wang2025opencua} advances this line via large-scale specialist fine-tuning, with 32B and 72B variants achieving strong grounding and navigation performance across diverse OS domains. DeepMiner-Mano-7B~\citep{fu2025mano} explores compact, parameter-efficient specialist architectures, balancing model scale and precise UI control.
Besides task-specialized models, generalist vision-language models (VLMs) and large language models (LLMs) have also been widely adapted for GUI tasks. The Qwen3-VL-32B family~\citep{bai2025qwen3} provides both standard instruction-following and explicit chain-of-thought reasoning variants, leveraging advanced visual understanding for screen interpretation. Claude-Sonnet-4.5~\citep{anthropic20253} and O3~\citep{openai2025introducing} represent state-of-the-art generalist backbones, employing long-context modeling and chain-of-thought to perform zero/few-shot GUI control. Although these generalist models deliver strong baseline reasoning, their performance in complex dynamic environments is still limited by insufficient grounding precision and strict step-budget constraints.

\subsection{Multi-Agent and Orchestrated Frameworks} 
Multi-agent paradigms outperform monolithic approaches in long-horizon tasks by leveraging modular decomposition and collaboration~\cite{cheng2024seeclick,wu2024atlas,zhang2025ufo2,zhao2025pyvision,li2025screenspot,qiu2025alita,wu2024autogen,agashe2024agent}. Representative works are as follows.
CoAct-1 \cite{song2025coact} introduces a dynamic orchestrator to allocate tasks between a GUI operator and a programmer agent, enabling hybrid graphical/code execution and achieving SOTA on OSWorld.
Agent S3~\cite{gonzalez2025unreasonable} enhances multi-agent coordination with refined global planning and constraint handling, improving success rates in multi-application workflows.
UiPath~\cite{cristescu2025ui} a hierarchical ReAct-style agent that employs an LLM-based planner to reason about task goals and generate action sequences, a visual grounder combining UI-TARS-7B with UiPath's computer vision model to map abstract actions to precise screen coordinates, and task/action reviewer components that close the execution feedback loop to trigger dynamic replanning when needed.
GTA-1 \cite{yang2025gta1} uses step-wise action sampling and a multimodal LLM judge to select optimal trajectories, improving robustness in complex GUI scenarios.
Jedi-7B optimizes step-wise decision-making via lightweight agentic fine-tuning, achieving efficient action selection with lower computational overhead.

%% file: tex/method.tex
\begin{figure*}[t]
  \centering
\includegraphics[width=0.98\textwidth]{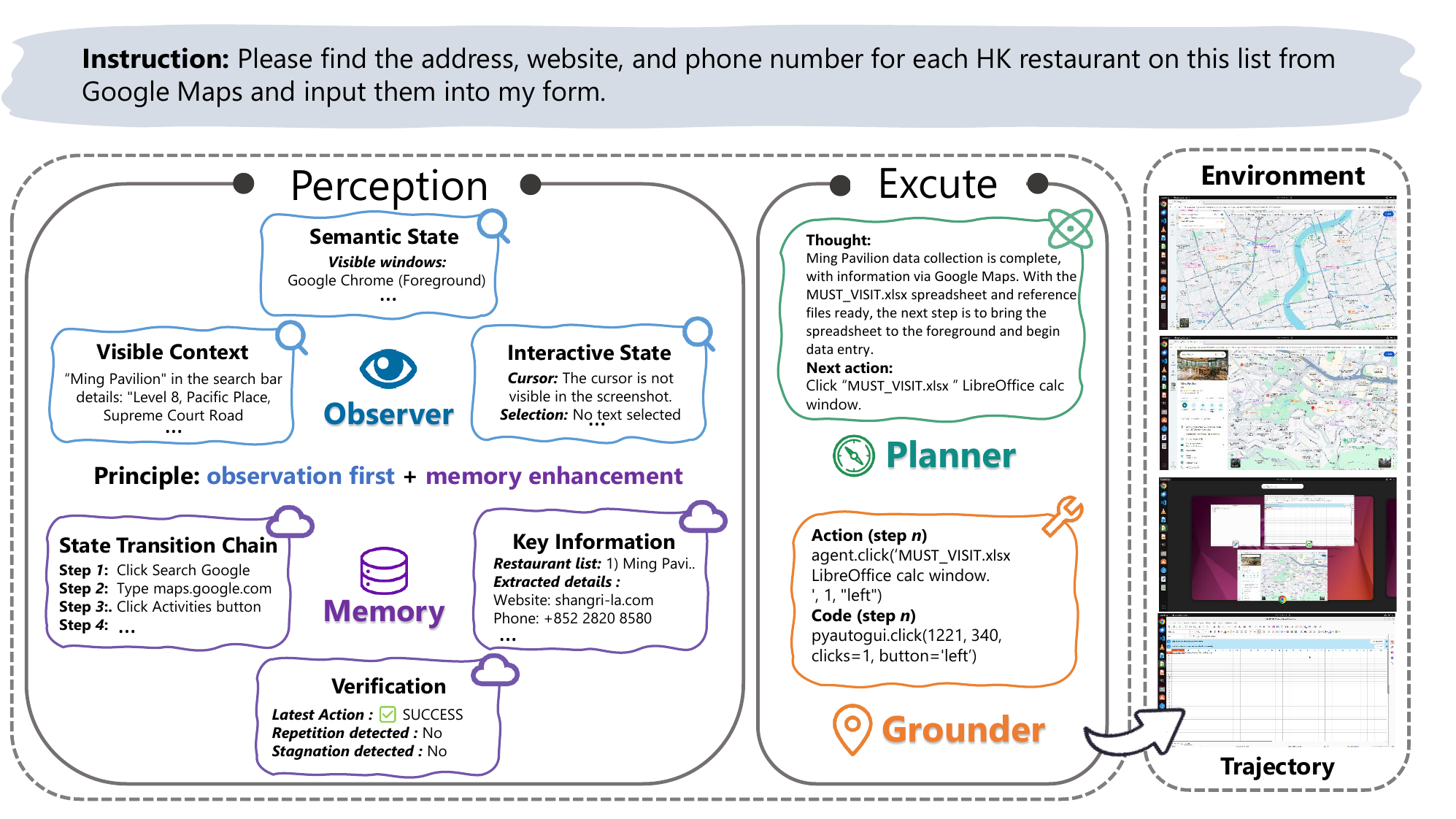}
  \caption{Detailed workflow of the proposed MGA framework on a representative \textit{\textbf{workflow}} domain task within OSWorld, illustrating the internal data flow among the Observer, Memory Agent, Planner, and Grounding pipeline.}
  \label{fig:lty-flow}
\end{figure*}

In this work, we introduce MGA, characterizing GUI interaction tasks as a closed loop dynamic system operating in discrete time steps.
As illustrated in Fig.~\ref{fig:lty-flow}, MGA integrates four cascaded, mutually complementary components that form a complete perception decision execution memory cycle: an Observation Agent, a Memory Agent, a Planning Agent, and a Grounding Agent.These components collaborate within a framework of clear data flows and functional boundaries, achieving unbiased perception, reliable memory, coherent planning, and accurate execution. The detailed prompts used in our work are shown in Appendix D.

\subsection{Problem Formulation}
Formally, we define the environment state at discrete time step \( t \) as a tuple that integrates current visual input, structured observation, and historical validated memory, capturing all critical information required for the agent’s decision making:
\begin{equation}
E_t = \big(I_t,\; Z_t,\; M_{t-1}\big),
\end{equation}
where: \( I_t \) denotes the raw screenshot of the GUI interface at step \( t \), serving as the fundamental input for perception; \( Z_t \) is the task agnostic spatial–semantic representation extracted by the Observation Agent from \( I_t \) (detailed in Section \ref{Observation}); \( M_{t-1} \) is the memory state distilled by the Memory Agent at step \( t-1 \) (detailed in Section \ref{memory}).

\subsection{Observation Agent}
\label{Observation}

Although many existing studies \cite{yang2025gta1,song2025coact} feed the raw screenshot $I_t$ to the Planning Agent, its internal attention mechanism is inherently decision driven, meaning it is heavily biased toward locating actionable interface elements to generate immediate actions. However, in complex GUI tasks, non-actionable information, such as dense text descriptions, read-only data content, or system state feedback, is indispensable for comprehending and verifying task objectives~\cite{li2025screenspot,yao2023react,shinn2023reflexion}. Relying solely on the planner to process raw screenshots inevitably leads to the neglect of this critical background semantics due to "action bias." Therefore, we introduce the Observation Agent as an independent, task agnostic perception module. Its core objective is to decouple perception from decision making.

By exhaustively extracting all visual semantic content, including pure text descriptions and topological structures on the screen, it provides the planner with a complete, unbiased global state reference, fundamentally mitigating the planner's inherent blind spots in information extraction.

Specifically, at time step $t$, given a GUI screenshot $I_t \in \mathcal{I}$, the observation function $\mathcal{O}$ leverages a vision language model to map the visual input to a state representation $Z_t$:
\begin{equation}
Z_t = \mathcal{O}(I_t) = \langle W_t, T_t, S_t \rangle,
\end{equation}
where the observation $Z_t$ is explicitly decoupled into three complementary dimensions:

\textbf{Application and Window Topology ($W_t$):} To determine the actual actionable area of the current system, we extract the global window structure of the interface. The window topology is defined as a set of active window states:
\begin{equation}
    W_t = \{(v_i, \phi_i, \mu_i)\}_{i=1}^{N_w},
\end{equation}
where $v_i$ represents the semantic identifier of the $i$-th window, $\phi_i \in \{0, 1\}$ is a boolean indicator denoting whether the window is currently in the foreground and has focus, and $\mu_i$ characterizes structural constraints such as modal dialogs or system-level overlays that block interaction.

\textbf{Visible Semantic Content ($T_t$):} To capture critical non-actionable descriptions and construct a complete semantic context, we extract an exhaustive inventory of all visible text on the screen along with their corresponding UI functional attributes. The semantic content is formulated as:
\begin{equation}
T_t = \{(x_j, c_j)\}_{j=1}^{N_t},
\end{equation}
where $x_j$ is the extracted text sequence, and $c_j \in \mathcal{C}$ represents the UI functional category to which the text belongs. This step strictly records the global semantic information of the interface without being constrained by current short-term operational goals.

\textbf{Dynamic Interaction State ($S_t$):} Since GUIs involve state transitions and transient responses, we record the dynamic interactive attributes of the interface through a state tuple:
\begin{equation}
S_t = \langle \kappa_t, \sigma_t, \rho_t, \tau_t \rangle,
\end{equation}
where $\kappa_t$ denotes the current element holding keyboard focus, $\sigma_t$ represents the currently selected text or regional range, $\rho_t$ encodes the relative scroll position of the current view within the global document, and $\tau_t$ captures transient temporal overlays. This dimension is crucial for aligning the agent's internal state with the actual responsiveness of the GUI.

\subsection{Memory Agent}
\label{memory}

To address the computational overload and logical chaos caused by directly stacking raw historical contexts, and to overcome the "loop" defect of traditional GUI agents that blindly trust instruction outputs~\cite{favreau2026multi, chen2024automanual}, we distill loose historical trajectories into an append-only state transition chain governed by mandatory validation functions.

Formally, at time step $t$, regarding the action $a_{t-1}$ executed in the previous round, the Memory Agent receives the historical memory $M_{t-1}$ and the dual frame screen captures $(I_{t-1}, I_t)$ before and after the operation. Through the memory update function $\mathcal{M}$, it outputs the structured validated memory $M_t$:
\begin{equation}
M_t = \mathcal{M}(M_{t-1}, a_{t-1}, I_{t-1}, I_t) = \langle V_t, H_t, E_t \rangle.
\end{equation}

The specific mechanisms and design objectives of each component are detailed as follows:

\textbf{(1) Dual Frame Visual Validation ($V_t$):} To strictly verify the real physical effects of the executed action, we introduce a dual frame visual comparison mechanism (Before/After Validation). Instead of simply recording action logs, this function compares the visual differences on the screen before and after the execution of $a_{t-1}$ to output a definite validation status:
\begin{equation}
V_t = f_{val}(a_{t-1}, I_{t-1}, I_t) \in \{\text{Success}, \text{Failure}, \text{Uncertain}\}.
\end{equation}

An action is deemed \text{Success} only when the target element exhibits a genuine visual state change. For operations judged as \text{Failure}, the system explicitly extracts the specific elements that failed to change, thereby physically blocking the agent's action hallucination.

\textbf{(2) Append-only State Transition Chain ($H_t$):} To avoid the contextual chaos brought by lengthy raw sequences, the Memory Agent deposits only the validated state increments ($\Delta S_t$) into the memory chain. The state chain follows a strict principle:
\begin{equation}
H_t = H_{t-1} \oplus \langle a_{t-1}, V_t, \Delta S_t \rangle,
\end{equation}
where $\oplus$ denotes the sequence concatenation operation. This fact-based memory construction approach replaces raw trajectories with structured high-level summaries, providing a verified and immutable contextual baseline for the downstream planner.

\textbf{(3) Anomaly Interceptor ($E_t$):} To equip the system with the critical capabilities of active error correction and loop escape, the memory mechanism embeds a Pattern Monitor acting as a real-time anomaly interceptor. This monitor actively detects behavioral pattern degradation by comparing the current action result with the historical records in $H_t$:
\begin{equation}
E_t = \begin{cases} 
\text{Repeated Error}, & \text{if } a_{t-1} \in H_{t-k} \text{ and } V_t = \text{Failure} \\ 
\text{Stagnant State}, & \text{if } I_t \approx I_{t-n} \text{ for } n \ge N_{threshold} \\ 
\text{None}, & \text{otherwise}. 
\end{cases}
\end{equation}

Once repetitive invalid operations on the same target (\text{Repeated Error}) or continuous physical stagnation of the interface state (\text{Stagnant State}) are detected, $E_t$ immediately triggers hard error alerts. These alerts are directly injected into the planning prompt for the next step, forcibly interrupting the model's current behavioral inertia and compelling it to re-evaluate the state and alter its exploration strategy~\cite{packer2023memgpt, madaan2023self, yao2023react}.

\subsection{Planning Agent}

The Planning Agent serves as the central cognitive hub of the system, responsible for converting high level user instructions into a coherent, executable sequence of actions. To address the computational overload and logical hallucination inherent in conventional long chain rollouts, our planner operates in a strict \emph{step-wise mode} with memory and observation. It evaluates the current environment and accumulated facts to produce an actionable decision without needing to replay the entire raw historical trajectory.

Formally, at step $t$, the planning function $\mathcal{P}$ conditions on a compact tuple to generate the intermediate reasoning trace (Thought, $th_t$) and the explicit next action ($a_t$):
\begin{equation}
\langle th_t, A_t \rangle = \mathcal{P}(Ins, I_t, Z_t, M_{t-1}),
\end{equation}
where $Ins$ is the user instruction, $I_t$ is the current raw screenshot, $Z_t$ is the structured, unbiased semantic observation provided by the Observation Agent, and $M_{t-1}$ is the append-only, validated state transition chain retrieved from the Memory Agent.

The planner's cognitive process is systematically decoupled into two sequential stages:~\textbf{1. High level Reasoning ($th_t$):} Before committing to a physical operation, the planner synthesizes the multi modal inputs to infer the logical next step. It evaluates the gap between the task agnostic current state ($Z_t$) and the user intent ($Ins$). Crucially, it consults the anomaly interceptor within the memory ($M_{t-1}$) to ensure the proposed strategy does not repeat previously failed attempts. \textbf{2. Action Specification ($a_t$):} Grounded in the reasoning trace $th_t$, the planner selects a concrete next action $a_t \in \mathcal{A}$. The action is formulated in a structured natural language or programmatic format, which is subsequently passed to the grounding module for physical execution.

\textbf{Action Space ($\mathcal{A}$):} To ensure versatility across diverse GUI environments, the action space $\mathcal{A}$ encompasses two distinct paradigms. First, it includes a comprehensive suite of atomic GUI operations grounded via executable primitives (e.g., \texttt{click}, \texttt{type}, \texttt{hotkey}). Second, it incorporates a specialized \texttt{code} action, empowering the agent to generate and execute Python scripts on the fly. This dual action space allows the planner to seamlessly switch between visual spatial UI interactions and complex, system level programmatic logic. The exhaustive definition and parameters of the action space $\mathcal{A}$ are detailed in the Appendix A.

\subsection{GUI Grounding Agent}

The GUI Grounding Agent transforms planner decisions into executable low level interactions. 
It receives the planner’s action $A_t$ specification, 
and maps them to precise screen coordinates or element references. 

Formally, the grounding agent outputs a pair:
\begin{equation}
a_t = (op, p) = G(A_t) ,
\end{equation}
where $op$ is a GUI primitive (click, double\_click, right\_click, type, hotkey, scroll) 
and $p$ is either a 2D coordinate or an element ID.
The grounding process unfolds in three stages:
\begin{enumerate}
    \item \textbf{Action parsing}. Interpret planner intent (e.g., ``Click Media'' $\to$ \texttt{click} operation). 
    \item \textbf{Element localization}. Resolve the intended target by consulting $Z_t$ (spatial layout + semantic roles). 
    \item \textbf{Execution binding}. Translate the operation into executable code, 
    such as \texttt{pyautogui.click()}.
\end{enumerate}

After execution, the grounding agent updates the environment state by reflecting the effect of the action 
(e.g., the Media drop down menu opens), which is then captured in the next observation $I_{t+1}$. 
This closes the loop between perception, planning, grounding, and memory, 
allowing the agent to iteratively refine its trajectory while remaining robust to misalignment and redundancy.

%% file: tex/exp.tex
\renewcommand{\arraystretch}{1.0}

\renewcommand{\arraystretch}{1.0}
\begin{table*}[!t]
    \centering
    \caption{
    We present a comparison of different categories of Frameworks on the OSWorld dataset. Reported values denote grounding accuracy (\%) across multiple domains, including Office, Daily, Professional, Operating System (OS), Workflow (Multi-Application), and the overall average (Overall). For each domain, ``Avg'' denotes the average score across its sub-tasks. The final overall averages are highlighted in bold. \textbf{Params (relative)} indicates the approximate model composition of each framework.
    }
    \begin{tabular}{lcccccccc}
        \toprule
        \textbf{Framwork} & \textbf{Steps}
        & \textbf{Office} & \textbf{Daily} & \textbf{Professional} 
        & \textbf{OS} & \textbf{Workflow} & \textbf{Overall} \\
        \midrule

        \multicolumn{8}{c}{\textbf{General Models \& Specialist Model}} \\
        \midrule
        O3                        & 50                 & 9.4    & 20.5   & 30.6   & 37.5  & 11.8  & 17.1 \\
        OpenCUA-32b               & 50                          & 29.9   & 40.9   & 65.3   & 60.9  & 14.6  & 34.1 \\
        OpenCUA-72B               & 100                        & 44.7  & 49.9 & 72.5  & 61.1 & 22.1 & 44.9 \\
        UI-TARS                   & 100                   & 50.4  & 55.6 & 51.0  & 41.6 & 14.6 & 41.8 \\
        UI-TARS-2                 & 100                & 61.1  & 62.1  & 61.2  & 41.6 & 34.1 & 53.1 \\
        DeepMiner-Mano-7B         & 100                         & 39.2  & 44.8  & 73.4  & 50.0 & 17.2 & 40.1 \\
        Claude-Sonnet-4.5         & 50              & 62.4 & 57.6 & 63.2 & 70.8 & 47.0 & 58.1 \\
        Qwen3-VL-32B-Instruct     & 50                          & -- & -- & -- & -- & -- & 32.4 \\
        Qwen3-VL-32B-Thinking     & 50                          & -- & -- & -- & -- & -- & 41.0 \\

        \midrule
        \multicolumn{8}{c}{\textbf{Agentic Framework}} \\
        \midrule
        Agent S3 w/ GPT-5-Mini    & 50         & 54.6 & 46.6 & 44.9 & 62.5 & 37.0 & 47.5 \\
        UiPath w/ GPT-5           & 50              & 49.5 & 62.1 & 71.4 & 73.9 & 37.3 & 53.7 \\
        Jedi-7B w/ o3             & 50            & 47.0  & 62.1  & 69.4  & 54.2  & 34.9  & 50.6 \\
        CoAct-1 w/ GPT-5          & 100            & 62.9  & 57.9  & 71.4  & 75.0 & 47.8 & 59.9 \\
        CoAct-1 w/ GPT-5          & 150              & 62.9  & 61.7  & 71.4  & 75.0 & 47.8 & 60.8 \\
        GTA1-7b w/ o3             & 50            & 46.1   & 44.9   & 77.6   & 58.3  & 37.1  & 48.6 \\
        GTA1-7b w/ o3             & 100           & 54.6   & 60.3   & 61.2   & 62.5  & 38.3  & 53.1 \\
        GTA1-7b w/ GPT-5          & 50           & -  & - & - & - & - & 61.0 \\
        OS-SYMPHONY w/ Qwen3-VL-32B  &50            &40.9 & 53.5 &75.1 &58.3 &31.2 &46.9 \\
        OS-SYMPHONY w/ GPT5-Mini  & 50        &58.2 &61.4 &75.0 &73.7 &47.4 &58.1\\
        OS-SYMPHONY w/ GPT-5      & 50              &\textbf{64.9} &61.2 &69.2 &75.0 &\textbf{54.9} &63.6\\
        \midrule[1pt]
        \multicolumn{8}{c}{\textbf{Our Methods: MGA}} \\
        \midrule
        MGA w/ o3                 & 50 & 49.0 & 62.8 & 70.6 & 68.2 & 36.4 & 52.9\\
        MGA w/ GPT-5              & 50 & 64.7 & \textbf{64.4} & \textbf{85.4}& \textbf{87.5} & 47.7 & \textbf{64.7}\\
        \bottomrule
    \end{tabular}
    \label{tab:OSWorld_comparison}
\end{table*}

\subsection{Dataset}
We evaluate our method on OSWorld \cite{xie2024osworld}. OSWorld is an extensible, practical computer environment designed for multimodal agents. It contains 369 tasks based on real web and desktop applications, covering file I/O, terminal operations, cross-application workflows (e.g., workflows spanning browser, email client, office suite, code editor, and system settings), multimedia processing, and automation scripting, exhibiting substantial domain diversity and long-horizon, multi-step decision complexity. OSWorld natively supports Ubuntu operating systems and unifies cross-platform execution, multimodal input/output, and direct interaction with real third-party applications in a single evaluation framework which is unique among existing benchmarks. This makes OSWorld a standardized testbed for assessing the generalization, robustness, and scalability of general-purpose agents. The benchmark is categorized into five major domains:

\subsection{Implementation Detail}

All experiments were conducted in the standardized OSWorld evaluation environment. For each task, OSWorld automatically establishes the initial environment (e.g., opening specified applications or websites, downloading files to designated directories), after which we capture an initial screenshot $I_0$ and provide it together with the user instruction to our system and baselines. For evaluation, we adopt the rule-based evaluator provided by OSWorld. Internally, the evaluator is composed of 134 handcrafted atomic execution-based predicates, each corresponding to a verifiable outcome of the system. For a given task, the benchmark combines these atomic predicates into Boolean expressions using logical AND/OR operators. A “pass” might require conditions such as (file\_exported AND MD5\_matches) AND (email\_sent == True). 

We evaluate two variants of our proposed MGA method: 1) MGA w/ o3, both the Planner and Memory Agent are built on o3; 2) MGA w/ GPT-5, we use GPT-5 as the Planner and qwen3-8b as the Memory Agent for cost considerations, while retaining the same overall architecture and functional logic. In practice, we employ Qwen-2.5-VL-7b as the foundational vision-language model to instantiate the observation function $\mathcal{O}$. We fine-tune it on the GUICourse dataset to inject GUI-domain visual grounding and UI parsing capabilities. Grounding agent adopts UI-TARS to translate the planner’s natural language actions into concrete GUI operations such as clicks, inputs, or drags. 

\subsection{Baseline}
We compare against diverse state-of-the-art methods categorized into three groups, with each method evaluated at the reasoning steps specified in the table. All comparisons are conducted on the OSWorld dataset, with reported values representing grounding accuracy (\%) across multiple domains and the overall average.

\textbf{General Models \& Specialist Model.}
This group includes general-purpose and domain-specialized models: 1) O3 \cite{openai2025introducing}, a cornerstone general-purpose benchmark model for open-domain reasoning (50 steps); 2) opencua-32b \cite{wang2025opencua}, an open-source human-computer interaction model (50 steps); 3) OpenCUA-72B, an upgraded large-scale variant (100 steps); 4) UI-TARS and UI-TARS-2, optimized for UI reasoning and control (100 steps); 5) DeepMiner-Mano-7B, a medium-scale model for targeted task execution (100 steps); 6) Claude-Sonnet-4.5 \cite{anthropic20253}, featuring enhanced long-context modeling (50 steps); 7) Qwen3-VL-32B-Instruct and Qwen3-VL-32B-Thinking, multimodal models with only overall accuracy reported (50 steps).

\textbf{Agentic Framework.}
We benchmark against prominent agentic frameworks with varied configurations: 1) Agent S3 w/ GPT-5-Mini, focused on multimodal agency (50 steps); 2) UiPath~\cite{cristescu2025ui} w/ GPT-5, specialized in real-world automation and screen interaction (50 steps); 3) Jedi-7B w/ o3 \cite{xie2025scaling}, emphasizing native architectural performance (50 steps); 4) CoAct-1 w/ GPT-5 \cite{song2025coact}, tailored for multi-task coordination (100 and 150 steps); 5) GTA1-7b w/ o3 \cite{yang2025gta1}, a lightweight flexible execution system (50 and 100 steps); 6) GTA1-7b w/ GPT-5, an upgraded variant of GTA1-7b (50 steps); 7) OS-SYMPHONY series, with variants paired with Qwen3-VL-32B, GPT5-Mini, and GPT-5 (50 steps each).

\subsection{Performance}
We conduct a quantitative comparative analysis on the OSWorld dataset, comparing MGA with mainstream general models, specialist models, and SOTA agentic frameworks (see Table \ref{tab:OSWorld_comparison}). Key results are as follows:

\textbf{Operating System (OS)}: The OS domain (6.5\% of tasks) evaluates fundamental interactions with file systems and CLI-GUI coordination. MGA w/ GPT-5 achieves $87.5\%$ grounding accuracy within a strict 50-step limit, significantly outperforming all the SOTA model and framework. Notably, CoAct-1 requires 150 steps to reach its peak, while MGA achieves superior control with fewer interactions, demonstrating highly efficient system level trajectory planning.

\textbf{Office}: The Office domain (31.7\%) involves complex document editing (LibreOffice) where agents must intervene in existing, high density files. This requires deep semantic perception and long-range dependency management. MGA w/ GPT-5 reaches $64.7\%$ accuracy, matching the performance of OS-SYMPHONY  ($64.9\%$) and surpassing CoAct-1 ($62.1\%$). Crucially, MGA achieves SOTA-level results within a strict 50-step horizon, outperforming baselines that utilize 100 or even 150 steps.

\textbf{Daily:} Daily tasks (21.1\%) involve frequent transitions between disparate apps like Chrome, VLC, and Thunderbird. These "fragmented" scenarios often cause agents to lose context during window switching. MGA w/ GPT-5 achieves $64.4\%$, outperforming all baselines (avg. $62.1\%$). This validates robustness of MGA’s observation and memory in handling heterogeneous GUI interfaces and its ability to maintain state across rapid visual changes.

\textbf{Professional:} The Professional domain (13.3\%) tests fine-grained operations in specialized software like VS Code and GIMP, requiring spatial reasoning and domain expertise. MGA w/ GPT-5 is highly competitive at $85.4\%$ while GTA1-7b w/ o3 leads at $77.6\%$, proving that MGA's architectural design effectively parses high-precision instructions without requiring massive parameter scaling.

\textbf{Workflow (Multi-Application):} Cross-Context Orchestration. As the most complex scenario (27.4\%), the Multi-Application domain requires orchestrating data across multiple platforms (e.g., Web to Spreadsheet to Email). MGA w/ GPT-5 achieves $47.7\%$. Although it trails OS-Symphony ($54.9\%$), partly due to the reasoning bottlenecks of the underlying Qwen3-8B, it still outperforms generalist agents like UiPath ($37.3\%$) and Agent S3 ($37.0\%$) under constrained steps. This indicates that while high-level logic remains a challenge, MGA provides a more reliable foundation for cross-app data transfer. A more granular investigation into this performance gap, is provided in Section \ref{sec:Performance Comparison with OS-Symphony}.

Overall, MGA w/ GPT-5 achieves an average accuracy of ($64.7\%$), establishing a new SOTA among existing methods. It outperforms OS-Symphony ($63.6\%$) as well as other strong baselines, including GTA1-7b ($61.0\%$) with 100 steps and CoAct-1 ($60.8\%$) with 150 steps, setting a new Pareto frontier between accuracy and inference efficiency. These results demonstrate that our observation module explicitly captures structural and content information of documents, enabling precise semantic understanding of complex layouts. Meanwhile, the memory module maintains long-range decision context and effectively reduces trial-and-error frequency, promoting more first-time-right decisions. The average token cost and statistical overhead analysis are presented in Appendix C.

\begin{table}[!ht]
\centering
\caption{Performance–cost trade-offs in the \textbf{Workflow} domain on 50 steps. The results validate that MGA achieves competitive success rates (SR) at substantially lower costs than redundant expert-ensemble frameworks. Cost calculations account for active modules under each configuration (Planner for GPT-5 and Memory for qwen3-8b; Planner and Memory for GPT-5).}
\label{tab:cost_comparison}
\begin{tabular}{lcc} 
\toprule
\textbf{Framework} & \textbf{Workflow SR (\%)} & \textbf{Cost (\$)} \\
\midrule
OS-Symphony & 54.9 & 150 \\
MGA (\textit{w/ Qwen3 memory}) & 47.7 & 15 \\
MGA (\textit{w/ GPT-5 memory}) & \textbf{56.3} & \textbf{64} \\
\bottomrule
\end{tabular}
\end{table}

\subsection{Ablation on Memory Module}
\label{sec:Performance Comparison with OS-Symphony}

To investigate the role of memory module representation capacity in long-horizon interaction tasks, we conducted controlled comparative experiments on a 50-step workflow domain by varying only the model used for the memory module while keeping the rest of the architecture unchanged. As shown in Table~\ref{tab:BetterMemory}, when using Qwen3-8B as the memory module, our framework achieved a success rate of 47.7\%, which is lower than OS-Symphony that employs GPT-5 throughout its entire pipeline (54.9\%). However, after replacing the memory module with GPT-5, our method's performance improved significantly to 56.3\%, surpassing the baseline under the same step constraints. 
This controlled substitution reveals that performance differences do not primarily stem from architectural complexity, but from the quality of memory state abstraction during long-horizon interactions.

Qwen3-8B exhibits representational capacity limitations in long-horizon intent retention, critical state milestone extraction, and implicit error tracing. These limitations lead to information compression loss in trajectory summarization and semantic drift of task intent as interaction steps increase, making it difficult to maintain execution robustness comparable to OS-Symphony in 50-step long-horizon sequences.

After upgrading the memory module to GPT-5, our framework, with only a single GPT-5 memory unit and a single GPT-5 planner, surpassed OS-Symphony which relies on the collaboration of six GPT-5 modules. This performance reversal reveals the following core insights: 1) High-quality state abstraction and trajectory summarization can implicitly compensate for the explicit scheduling and verification overhead in multi-agent systems, maintaining context consistency without additional verification modules. 2) In computer-use agents, the impact of long-term memory fidelity on final decision-making performance is significantly greater than the complexity of modular engineering design. 3) A minimalist architecture combined with high-capacity memory models can achieve superior long-horizon task planning and execution robustness with significantly reduced inference overhead and system complexity.
Our framework retains only a minimal viable pipeline: a visual perception module (Observer, Qwen-VL-7B), a decision-making module (Planner, GPT-5), and a memory abstraction module (Memory), explicitly removing auxiliary scheduling, multi-module state verification, and complex coordination mechanisms. In the initial configuration, only the planner uses a high-capacity model. Detailed case studies on the iterative transformation of memory are provided in Appendix B.

\subsubsection{Performance-CostTrade-offs}
Notably, table~\ref{tab:cost_comparison} reports the success rates and operational costs under different model configurations.

\textbf{Efficiency Frontier:} When utilizing GPT-5 as the memory model, MGA achieves a $56.3\%$ success rate at a cost of $\$64$ per task. In comparison, OS-Symphony requires an expenditure of $\$150$ to reach a $54.9\%$ success rate. The data suggests that MGA’s coordinated "observation-memory" logic is more efficient for practical tasks than frameworks built on redundant expert model stacks.

\textbf{Cost Elasticity:} MGA demonstrates significant cost elasticity. By switching to a lightweight memory model (Qwen3), the cost per task drops to $\$15$—approximately $10\%$ of the OS-Symphony’s cost. Even in this high-efficiency configuration, MGA retains a success rate of $47.7\%$, validating the framework’s viability for large-scale deployment or resource-constrained environments.

\begin{table}[t]
    \centering
    \caption{
    Ablation study on core components. 
    \textbf{w/o obs} removes the structured observation $Z_t$; 
    \textbf{w/o memory} empties the distilled memory $\mathcal{M}_{t-1}$ (reverting to plain trajectory execution). 
    Reported values are grounding accuracy (\%) across domains.
    }
    \begin{tabular}{lcc}
        \toprule
        \textbf{Configuration} & \textbf{GPT-5} & \textbf{O3}  \\
        \midrule
        MGA (Full)               & 56.3  & 36.4  \\
        - w/o obs                & 46.0  & 35.1  \\
        - w/o memory             & 39.0  & 27.7  \\
        - w/o obs + w/o memory   & 36.8  & 15.8  \\
        \bottomrule
    \end{tabular}
    \label{tab:ablation_study}
\end{table}

\subsection{Ablation on Core Components}
To verify the necessity and synergistic effects of MGA's core components, we conduct ablations on the Workflow domain, with results summarized in Table \ref{tab:ablation_study}.

\textbf{Memory Module.} Removing the memory module (\textbf{w/o memory}, emptying $\mathcal{M}_{t-1}$) causes a sharp accuracy drop to 39.0\% (GPT-5) and 27.7\% (O3). This confirms the memory module’s core role: it distills historical trajectories to avoid context bloat and state forgetting, which is critical for long-horizon tasks.

\textbf{Observation-First Mechanism.} Eliminating the structured observation $Z_t$ (\textbf{w/o obs}) reduces accuracy to 46.0\% (GPT-5) and 35.1\% (O3). The observation module mitigates action bias by extracting complete semantic content and topology, thereby reducing interference from transient visual noise.

\textbf{Synergistic Effects.} Removing both components yields the worst performance (36.8\% for GPT-5, 15.8\% for O3). The significant drop demonstrates that $Z_t$ and $\mathcal{M}_{t-1}$ are highly synergistic: perception filters noise, and memory condenses semantics, together forming a robust state-tracking pipeline.

%% file: tex/conclution.tex
We introduce MGA, a lightweight agent paradigm that rethinks interaction pipelines by replacing monolithic context windows and bloated collaboration protocols with decoupled decision nodes anchored by a structured memory chain. The Observer First mechanism ensures unbiased, high-fidelity state perception by strictly separating factual screen reading from intent-driven planning, thereby mitigating visual hallucinations and action bias. Complementarily, the Structured Memory mechanism replaces verbose historical logs with an append-only, validation-gated state transition chain. By enforcing dual-frame visual comparison after each action, the system effectively intercepts error cascades, prevents behavioral stagnation, and enables proactive online correction. Empirical evaluations on OSWorld and diverse real-world applications confirm that MGA delivers competitive task success rates without relying on heavyweight toolsets or extensive context management. Our results demonstrate that architectural minimalism, grounded state tracking, and disciplined perception are sufficient to unlock reliable long-horizon GUI automation. While MGA significantly reduces cognitive load and inference latency, its reliance on baseline MLLM capabilities for complex visual grounding presents a natural boundary. Future work will explore adaptive memory compression strategies, cross-application state transfer, and tighter integration with low-level execution engines to further enhance robustness in highly dynamic, open-world environments. Ultimately, MGA provides a principled, efficient alternative to bloated agent frameworks, paving the way for more scalable, deployable, and human-aligned GUI automation systems.

%% file: appendix.tex
\clearpage
\appendix

\section{Action Abstraction and Parameterized Schema}

A core design challenge in GUI agents is defining an action abstraction that remains robust across long-horizon task execution, varying screen layouts, and ambiguous interface states. As shown in Table~\ref{tab:agent_action_space}, MGA adopts a discrete, parameterized action space \( \mathcal{A} \) that combines semantic GUI interactions with programmatic execution through a dedicated \texttt{code} action to enable robust desktop automation. We represent each action as a parameterized tuple \( a_t = (op, p) \), where $op$ denotes the action type and $p$denotes its associated arguments.

\textbf{Semantic-grounded GUI Actions.} Following prior work such as \cite{yang2025gta1}, the GUI action set includes atomic interaction primitives such as \texttt{click}, \texttt{drag\_and\_drop}, \texttt{type}, \texttt{scroll}, and \texttt{hotkey}. Each action is parameterized by a semantic target (e.g., a natural language description of a UI element) and optional action-specific arguments (e.g., input text, scroll direction, key combinations).

Instead of relying on fixed coordinates, MGA resolves interaction targets through a Visual Grounding Model at execution time. Given a semantic description, the grounding module predicts the corresponding screen location, enabling robustness to layout variation and resolution changes. Variants of actions (e.g., single vs.\ double click) are expressed through parameters, ensuring that each action remains atomic.

\textbf{Programmatic Execution via \texttt{code}.}  
In addition to GUI actions, the action space includes a \texttt{code} action that allows the agent to generate and execute Python scripts, with execution results returned to the environment. This action is parameterized by an executable code snippet, which operates directly on system-level resources (e.g., files, structured data). It provides a complementary interaction modality when GUI-based manipulation is inefficient or imprecise, such as in dense interfaces (e.g., spreadsheets) or background operations.

\textbf{Control and Termination Signals.}  
To support stable long-horizon execution, we introduce a small set of control signals. The \texttt{wait} action pauses execution to accommodate asynchronous system responses. The \texttt{done} and \texttt{fail} signals indicate successful completion and unsuccessful termination, respectively. These signals are not executable interactions but are included to explicitly manage task flow during rollout.

\renewcommand{\arraystretch}{1.0}
\begin{table*}[t]
    \centering
    \caption{Agent action space with parameterized action schema.}
    \label{tab:agent_action_space}
    
    \begin{tabularx}{\textwidth}{
        >{\raggedright\arraybackslash}p{2.4cm}
        >{\raggedright\arraybackslash}p{3.8cm}
        >{\raggedright\arraybackslash}p{3.4cm}
        >{\raggedright\arraybackslash}X}
        
        \toprule
        \textbf{Action} & \textbf{Parameters} & \textbf{Description} & \textbf{General Constraints} \\
        \midrule
        
        \multicolumn{4}{c}{\textbf{Semantic-grounded GUI Actions}} \\
        \midrule
        
        \textbf{click} & 
        \texttt{instruction: str}\newline
        \texttt{num\_clicks: int}\newline
        \texttt{button\_type: str}\newline
        \texttt{hold\_keys: list} & 
        Click on a semantically described UI element. & 
        Target elements are specified via semantic descriptions.\\
        \addlinespace[6pt]
        
        \textbf{drag\_and\_drop} & 
        \texttt{starting\_desc: str}\newline
        \texttt{ending\_desc: str}\newline
        \texttt{hold\_keys: list} & 
        Drag from one semantically described element to another. & 
        Both source and target are resolved via semantic grounding.\\
        \addlinespace[6pt]
        
        \textbf{type} & 
        \texttt{element\_desc: str}\newline
        \texttt{text: str}\newline
        \texttt{overwrite: bool}\newline
        \texttt{enter: bool} & 
        Type text into the specified UI element. & 
        Supports optional overwrite and submission behavior. \\ 
        \addlinespace[6pt]
        
        \textbf{hotkey} & 
        \texttt{keys: list} & 
        Execute a keyboard shortcut. & 
        Keys are provided as standard key combinations. \\
        \addlinespace[6pt]
        
        \textbf{scroll} & 
        \texttt{instruction: str}\newline
        \texttt{clicks: int}\newline
        \texttt{shift: bool} & 
        Scroll within a semantically specified region. & 
        Scroll direction and magnitude are parameterized. \\

        \midrule
        
        \multicolumn{4}{c}{\textbf{Programmatic Execution via \texttt{code}}} \\
        \midrule
        
        \textbf{code} & 
        \texttt{code: str} & 
        Execute a Python script for direct system-level operations. & 
        Enables precise manipulation of structured data and background resources. \\
           
        \midrule
        
        \multicolumn{4}{c}{\textbf{Control and Termination Signals}} \\
        \midrule
        
        \textbf{wait} & 
        \texttt{time: float} & 
        Pause execution for a specified duration. & 
         Handles asynchronous UI responses and latency. \\
        \addlinespace[6pt]
        
        \textbf{done} & 
        \texttt{return\_value: Optional} & 
        Signal successful task completion. & 
        Terminates execution and optionally returns results.\\
        \addlinespace[6pt]
        
        \textbf{fail} & 
        None & 
        Signal unsuccessful termination. & 
        Used when the task cannot be completed under the current trajectory. \\
        
        \bottomrule
    \end{tabularx}
\end{table*}

\begin{figure*}[t]
    \centering
    
    \begin{subfigure}[b]{0.48\textwidth}
        \centering
        \includegraphics[width=\textwidth]{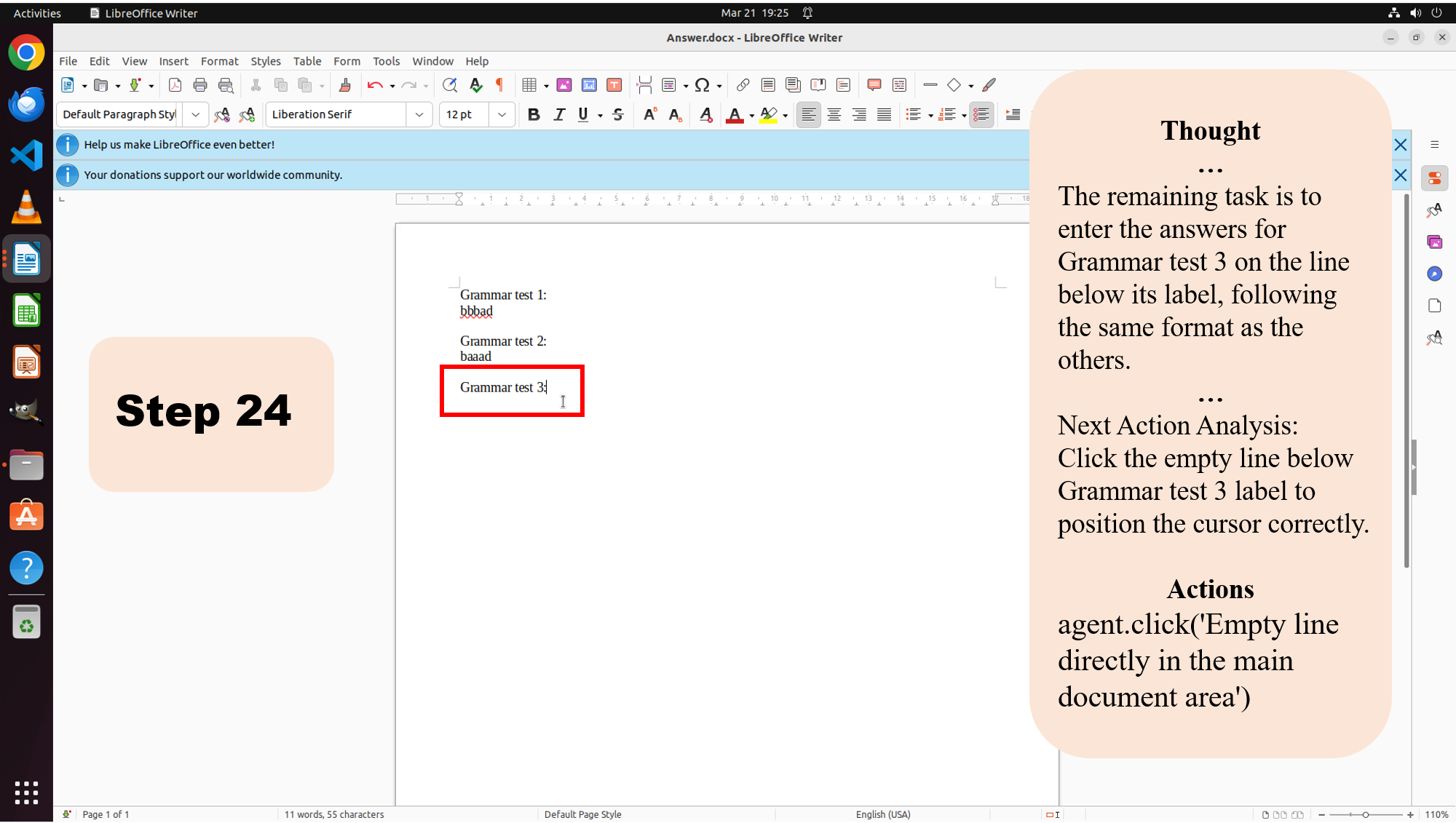} 
        \caption{Step 24: Ready to input answer}
        \label{fig:recovery_a}
    \end{subfigure}
    \hfill
    \begin{subfigure}[b]{0.48\textwidth}
        \centering
        \includegraphics[width=\textwidth]{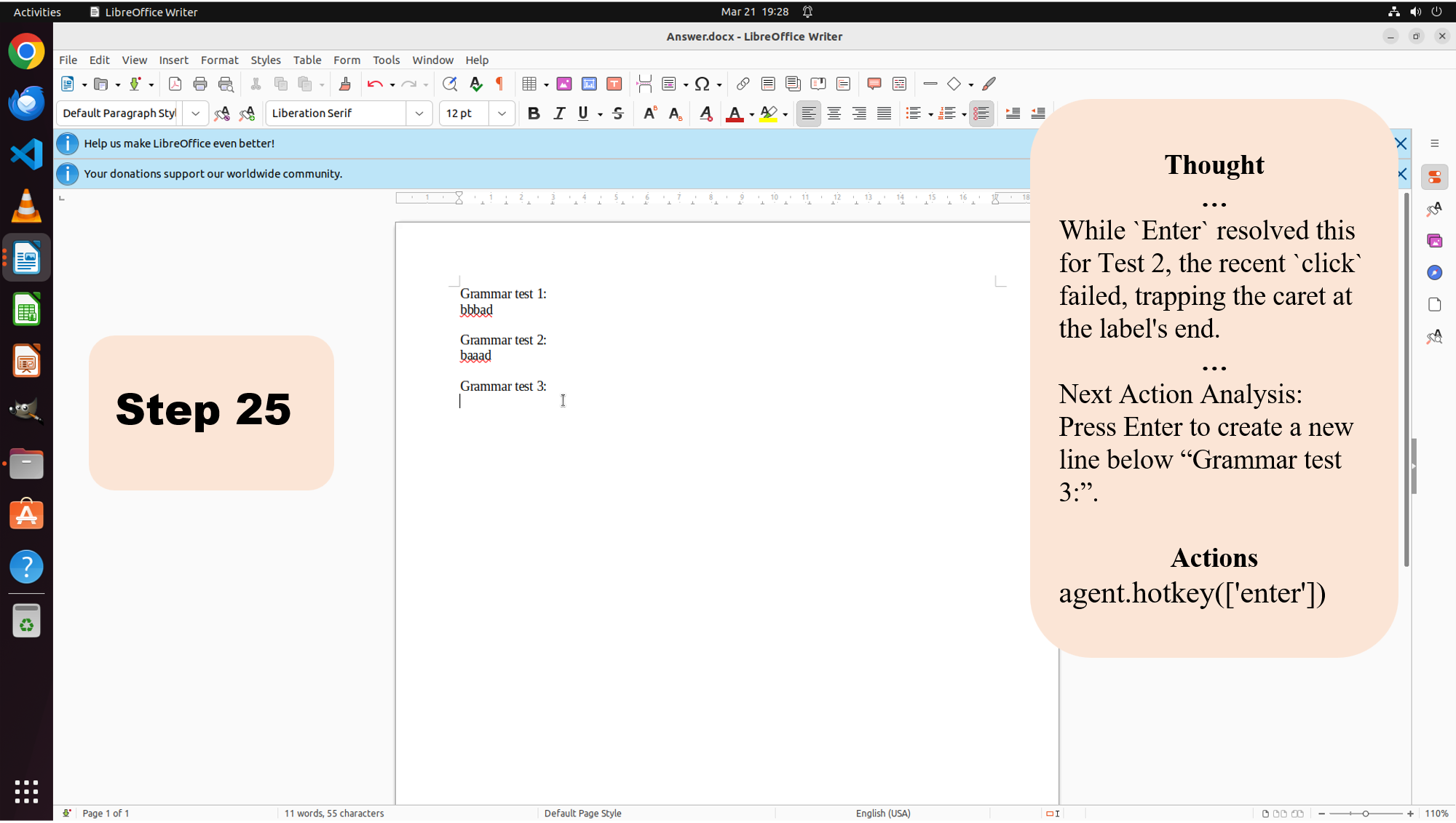} 
        \caption{Step 25: Stagnation (Enter key fails)}
        \label{fig:recovery_b}
    \end{subfigure}
    
    \vspace{1.5em} 

    \begin{subfigure}[b]{0.48\textwidth}
        \centering
        \includegraphics[width=\textwidth]{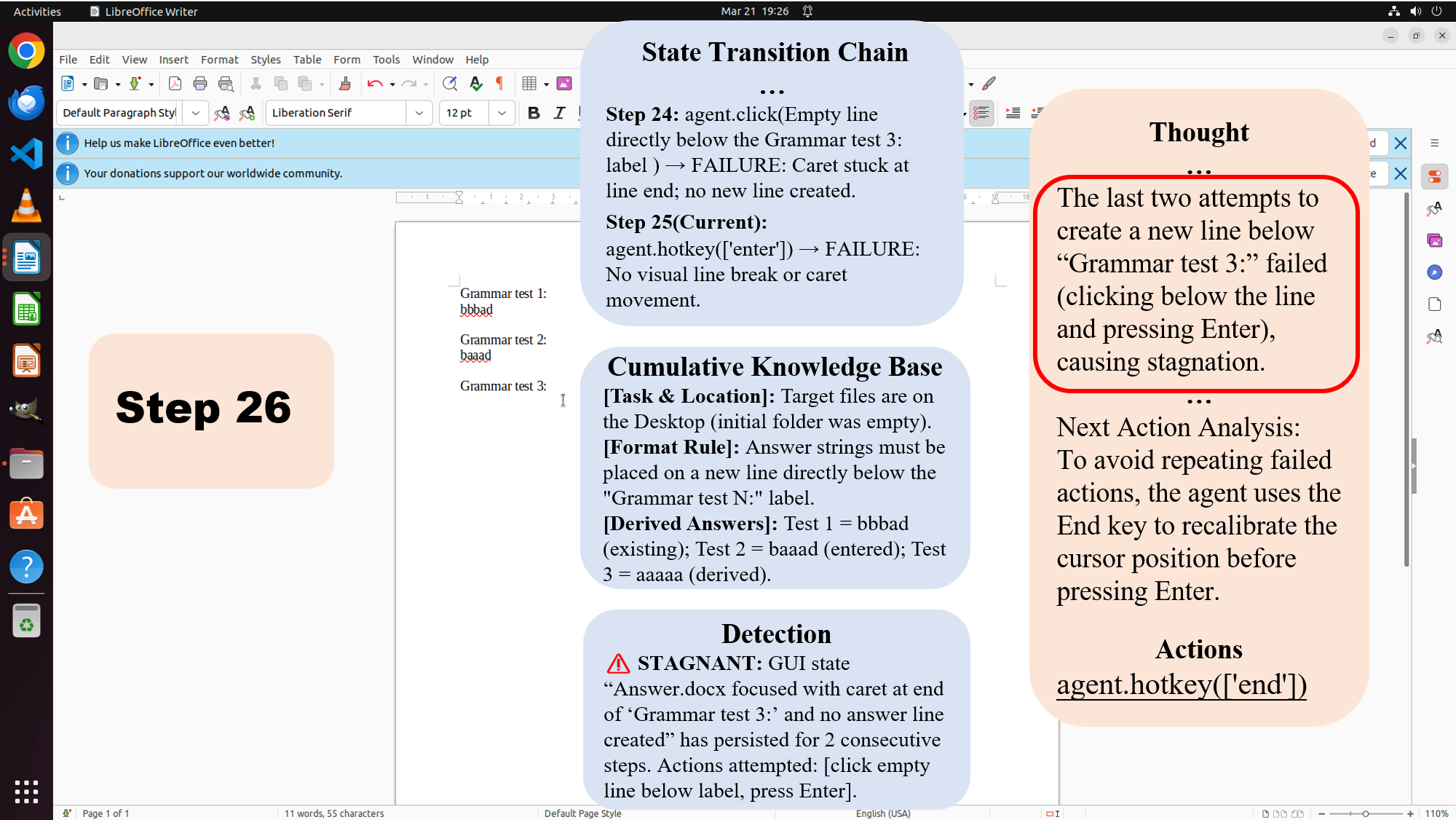} 
        \caption{Step 26: Memory triggers replanning trajectory}
        \label{fig:recovery_c}
    \end{subfigure}
    \hfill
    \begin{subfigure}[b]{0.48\textwidth}
        \centering
        \includegraphics[width=\textwidth]{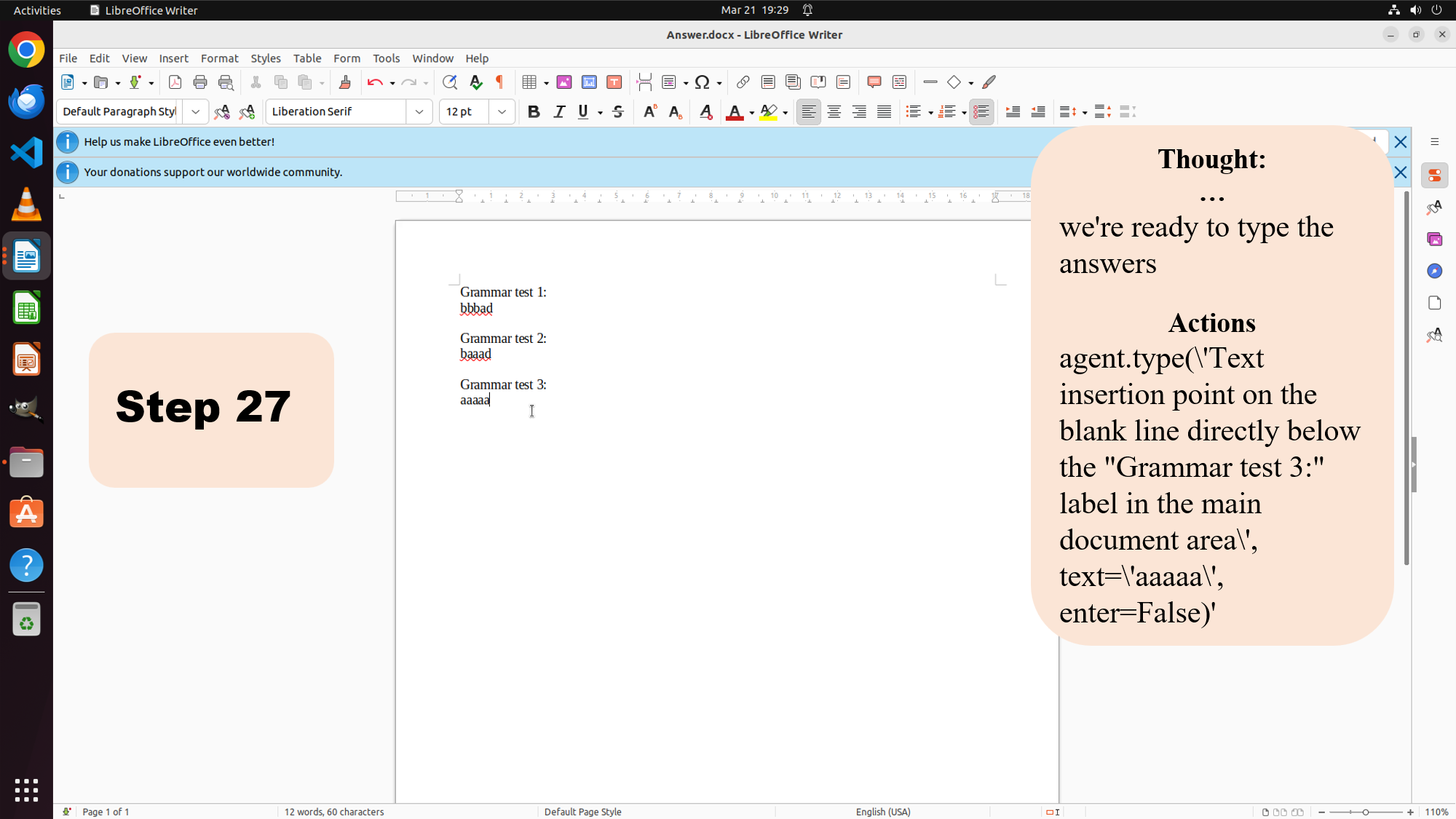} 
        \caption{Step 27: Recovery (Successfully typed 'aaaaa')}
        \label{fig:recovery_d}
    \end{subfigure}
    
    \caption{Visualizing the Anti-Stagnation Recovery process. (a) The agent prepares to input the answer. (b) When basic operations fail to create a new line due to UI ambiguity, a stagnation occurs. (c) The memory module detects the lack of visual state change and triggers a replanning trajectory. (d) The agent autonomously adapts by injecting an \texttt{End} hotkey to recalibrate the cursor, successfully completing the text insertion.}
    \label{fig:stagnation_recovery}
\end{figure*}

\renewcommand{\arraystretch}{1.0}
\begin{table*}[!t]
    \centering
    \caption{Timeline of memory evolution and agent strategy adjustment.}
    \label{tab:memory_timeline}

    \begin{tabularx}{\textwidth}{
        >{\raggedright\arraybackslash}p{3.2cm}
        >{\raggedright\arraybackslash}p{1.8cm}
        >{\raggedright\arraybackslash}X
        >{\raggedright\arraybackslash}X}
        
        \toprule
        \textbf{Phase} & \textbf{Steps} & \textbf{Memory State Update} & \textbf{Consequent Agent Action} \\
        \midrule
        
        \textbf{1. Knowledge Correction} & 
        1--8 & 
        Action verification detects ``Folder is empty'', correcting the knowledge base: files reside on the Desktop rather than the target folder. & 
        Aborts folder search and redirects grounding toward Desktop icons via the open dialog. \\
        \midrule
        
        \textbf{2. Working Memory} & 
        9--23 & 
        Aggregates fragmented options from scrolling interactions and stores derived answers (e.g., Test 2 = ``baaad'', Test 3 = ``aaaaa''). & 
        Returns to Answer.docx using pre-computed semantic strings, avoiding redundant copy-paste operations. \\
        \midrule
        
        \textbf{3. Anti-Stagnation} & 
        24--27 & 
        Stagnation detection triggers after repeated \texttt{click} and \texttt{enter} fail to produce visible changes. & 
        Terminates ineffective retries and applies a fallback strategy (\texttt{End} hotkey) to reset cursor position. \\
        \midrule
        
        \textbf{4. Loop Avoidance} & 
        28--33 & 
        Action verification returns \texttt{UNCERTAIN} due to lack of visual feedback during save operations and flags repeated hotkey usage. & 
        Prioritizes global task completion over local uncertainty and safely terminates via \texttt{done()}. \\
        
        \bottomrule
    \end{tabularx}
\end{table*}

\section{Case Study: Dynamic Evolution of Agent Memory}
\label{app:case_study_memory}

To demonstrate the efficacy of our memory module in long-horizon tasks, we present a detailed case study of the agent executing a multi-document information extraction and formatting task. 

\textbf{Task Description:} The agent is instructed to open the ``Grammar test'' folder, extract the multiple-choice answers for Test 2 and Test 3, and write them into an \texttt{Answer.docx} file, strictly following the formatting of Test 1 (e.g., ``\texttt{bbbad}''). 

Over the course of 33 steps, the agent's memory module dynamically evolves, proving crucial for error recovery, cross-context reasoning, and infinite loop avoidance. The evolution can be divided into four distinct phases (Table.~\ref{tab:memory_timeline}):

\textbf{Phase 1: Environmental Exploration and Knowledge Correction (Steps 1--8)}

Initially, based on the user's prompt, the agent assumes the target files are inside the ``Grammar test'' directory on the desktop. Upon opening the folder (Step 1), the visual observation reveals a ``Folder is Empty'' message. At this juncture, the \textit{Latest Action Verification} flags the unexpected state, prompting an immediate update to the \textit{Cumulative Knowledge Base}: \textit{``The 'Grammar test' folder is empty; the relevant files are on the Desktop.''} This dynamic correction prevents the agent from hallucinating or endlessly searching within the empty directory, seamlessly redirecting its attention to the desktop files.

\textbf{Phase 2: Cross-Document Working Memory (Steps 9--23)}

In this phase, the agent must retain the target format from \texttt{Answer.docx} while navigating other documents. During the exploration of \texttt{Grammar test 2.docx} and \texttt{Grammar test 3.docx}, the agent utilizes scrolling actions to gather fragmented visual information (individual test questions). The \textit{Cumulative Knowledge Base} acts as a working memory, successfully abstracting raw visual inputs into condensed logical strings (e.g., logging \textit{``Derived Test 2 answer string: 'baaad'''}). This prevents the agent from forgetting early answers while reading subsequent pages.

\textbf{Phase 3: Anti-Stagnation Recovery (Steps 24--27)}

A critical bottleneck occurs when the agent attempts to insert the derived answers into \texttt{Answer.docx}. The agent tries to place the cursor on an empty line below the ``Grammar test 3:'' label using a mouse click (Step 24, Figure~\ref{fig:recovery_a} ) and the \texttt{Enter} key (Step 25, Figure~\ref{fig:recovery_b}), both of which fail to produce a visible line break. 

At Step 26 (Figure~\ref{fig:recovery_c}), \textit{Repetition \& Stagnation Detection} memory is triggered, throwing a critical warning: \texttt{STAGNANT: GUI state has persisted for 2 consecutive steps}. 

Triggered by this explicit warning, the agent aborts the failing trajectory. It reasons that the cursor position is ambiguous and autonomously falls back to a more robust strategy: pressing the \texttt{End} key to ensure the cursor is at the absolute end of the line before pressing \texttt{Enter} again. This self-healing capability breaks the deadlock without human intervention, and thus finally make the phase success in step 27(Figure~\ref{fig:recovery_d}).

\textbf{Phase 4: Goal Prioritization and Infinite Loop Avoidance (Steps 28--33)}

In the final phase, the agent attempts to save the document using \texttt{Ctrl+S} (Step 28) and the UI save button (Step 29). However, due to LibreOffice's silent saving mechanism, the \textit{Latest Action Verification} returns an \texttt{UNCERTAIN} state, as no visual confirmation dialog appears. 

By Step 33, the memory module flags a \texttt{REPEATED} warning for the save hotkey. Instead of getting stuck in an infinite loop of attempting to verify the save, the agent consults its \textit{Cumulative Knowledge Base}, verifying that the core objective (writing ``baaad'' and ``aaaaa'' in the correct format) is visually confirmed. Prioritizing the global task completion over local sub-step uncertainty, the agent gracefully terminates the process by calling \texttt{agent.done()}.

\section{System-Level Resource and Inference Analysis}
\label{appendix:efficiency_analysis}

In this section, we provide a quantitative evaluation of the MGA framework's efficiency. By analyzing inference overhead, modular resource distribution, and operational costs, we demonstrate how our first-principles design addresses the dual bottlenecks of historical interference and system redundancy.

\subsection{Inference Efficiency and Context Management}

Table~\ref{tab:token_comparsion_all_task} compares the inference overhead of MGA against OS-Symphony. In contrast to mainstream frameworks that rely on an ensemble of expert agents to manage complexity, MGA utilizes a streamlined single-step execution logic.

\textbf{Exploration Cost and Global Efficiency: }Experimental results indicate that MGA’s average execution step count ($19.3$) is higher than the OS-Symphony ($15.2$). Admittedly, this minimalist design may require a higher number of exploration steps to complete certain complex tasks. However, due to MGA’s efficient distillation and compression of interaction history, its prompt token consumption per step is significantly lower than that of the OS-Symphony. Even with a higher step count, MGA’s total token consumption ($226.1$K) is only approximately $38\%$ of the OS-Symphony’s ($598.6$K), demonstrating the global economic advantage of a low-cost per-step design.

\begin{table*}[t]
\centering
\caption{Inference efficiency on OSWorld. MGA’s minimalist single-step logic results in a higher average step count for exploration; however, due to low per-step costs achieved through history distillation, total token consumption remains significantly lower than the OS-Symphony. Token statistics (in thousands, K) include prompt, completion, and total tokens averaged across all task categories.
}
\label{tab:token_comparsion_all_task}
\begin{tabular}{l c c c c} 
\toprule
\textbf{Framework} & \textbf{Avg. Step} & \textbf{Prompt Tokens} & \textbf{Completion Tokens} & \textbf{Total} \\
\midrule
OS-Symphony & 15.2 & 572.9K & 25.7K & 598.6K \\
MGA & 19.3 & 188.8K & 37.3K & 226.1K  \\
\bottomrule
\end{tabular}
\end{table*}

\subsection{Modular Design Rationale}

\begin{figure*}[t]
  \centering
  \includegraphics[width=0.5\linewidth]{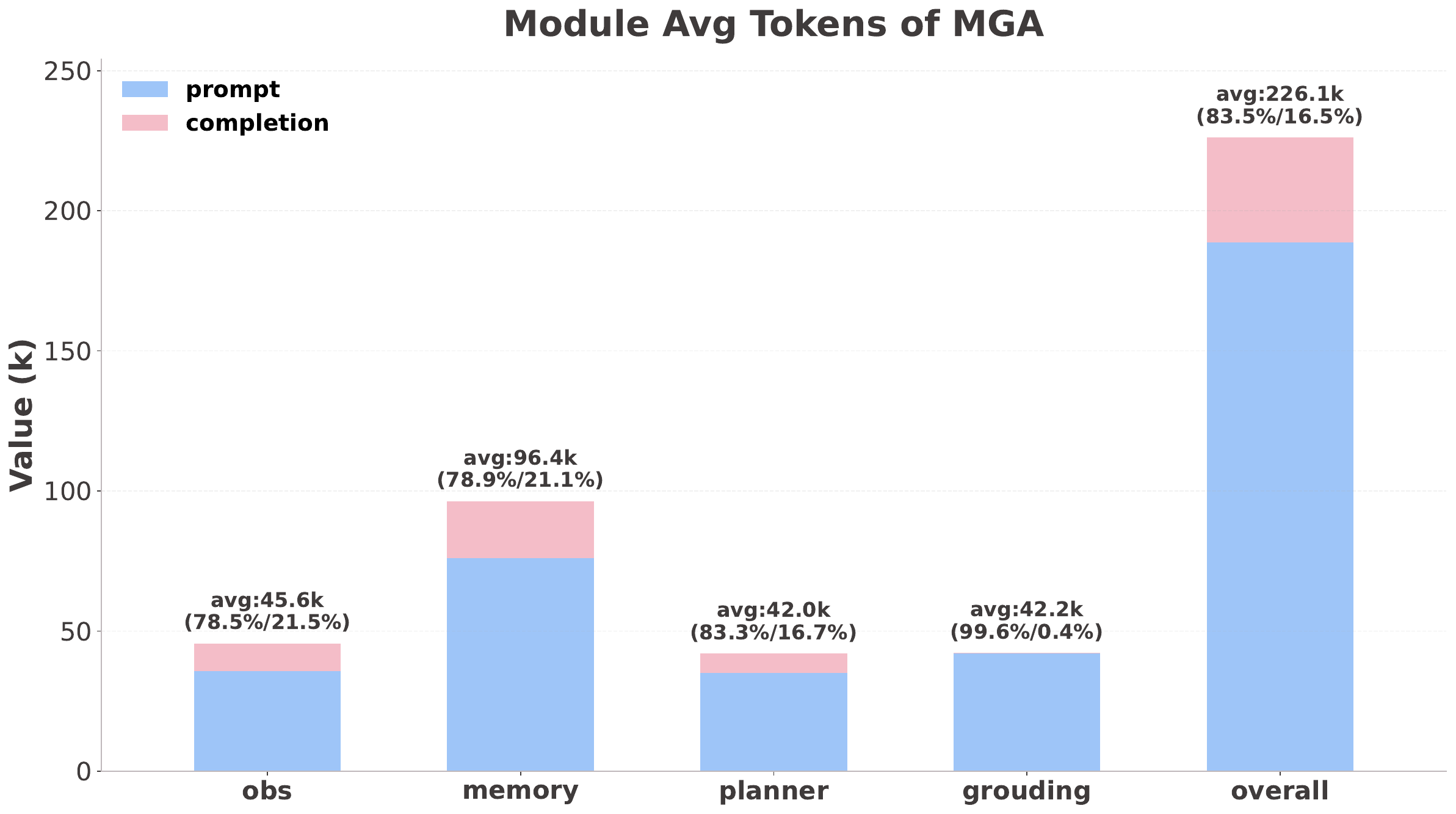}
  \caption{Token distribution across MGA modules. The dominant overhead of the Memory module (42.6\%) reflects its core role in the memory-driven architecture, while the Observer module provides essential perceptual granularity. Stacked bars represent average prompt and completion tokens (in thousands) for each module, with relative percentages indicating their respective proportions of total system overhead.}
  \label{fig:module-avg-tokens-mga}
\end{figure*}

Figure~\ref{fig:module-avg-tokens-mga} illustrates the distribution of token consumption across MGA’s modules. As a Memory-driven Agent, the Memory module accounts for the largest proportion of token overhead ($42.6\%$), as it performs the computationally intensive tasks of state validation and history compression.

\textbf{Indispensability of the Observer Module:} The Observer module’s consumption remains stable at $45.6$K tokens. This module is not designed for cost reduction; rather, it is an essential component that provides a significantly higher level of perceptual granularity than a standalone Planner. While the Observer introduces higher-dimensional state descriptions, the modular architecture ensures that this added detail does not compromise overall economy. MGA’s total operational cost remains substantially lower than that of the OS-Symphony.

\section{Prompt Design}
To ensure modular synergy and minimize context redundancy, we developed a set of role-specific system prompts based on the principle of functional isolation.

\textbf{Memory agent:} The Memory agent (Fig. \ref{fig:mem_prompt}) is tasked with maintaining the \textit{State Transition Chain}. Unlike naive history concatenation, our prompt enforces an \textbf{Append-Only} rule and a \textbf{Verification Logic Rule}. This forces the model to distill unprocessed interaction logs into verified state increments, which is the key driver behind the inference efficiency discussed in Section \ref{appendix:efficiency_analysis}.

\textbf{Observer:} The Observer's prompt (Fig. \ref{fig:obs_prompt}) is strictly constrained to report only visible GUI states. By explicitly prohibiting intent inference ("DO NOT INFER INTENT"), we decouple raw perception from subjective planning, thereby mitigating confirmation bias in complex UI environments.

\textbf{Planner: }The Planner (Fig. \ref{fig:plan_prompt}) operates on a coordinate-free Python API. By mandating functional element descriptions (e.g., "the Search bar") over numeric pixels, the prompt shifts the Planner's focus from low-level spatial grounding to high-level strategic reasoning. Furthermore, the mandatory \texttt{Thought-Action} template ensures that each decision is grounded in the distilled history provided by the Memory module.

\begin{figure*}[t]
\centering
\caption{MEMORY AGENT SYSTEM PROMPT}
\label{fig:mem_prompt}
\begin{tcolorbox}[
    enhanced,
    colback=white,
    colframe=gray,
    arc=0mm,
    boxrule=1.2pt,
    title=\textbf{\textsf{MEMORY AGENT SYSTEM PROMPT}},
    fonttitle=\bfseries\large,
    coltitle=white,
    attach boxed title to top center={yshift=-3mm},
    boxed title style={colback=gray, sharp corners},
    shadow={1.5mm}{-1.5mm}{0mm}{black!10!white} 
]
    \vspace{2mm}
    
    \begin{minipage}[t]{0.32\linewidth}
        \textbf{1. STATE TRANSITION CHAIN}
        \rule{\linewidth}{0.6pt} \\
        \footnotesize
        \begin{itemize}[leftmargin=1.2em, itemsep=0pt, topsep=2pt]
            \item \textbf{Narrative}: Single coherent chain.
            \item \textbf{Append-Only}: Lock previous history; append only latest verified delta.
            \item \textbf{Delta}: Explicitly record failures or "no visible effect" steps.
        \end{itemize}
    \end{minipage}
    \hfill
    \begin{minipage}[t]{0.32\linewidth}
        \textbf{2. CUMULATIVE KNOWLEDGE}
        \rule{\linewidth}{0.6pt} \\
        \footnotesize
        \begin{itemize}[leftmargin=1.2em, itemsep=0pt, topsep=2pt]
            \item \textbf{Extraction}: Save specific IDs, document rules, or key requirements.
            \item \textbf{Goal Tracking}: Record high-level goal updates discovered in GUI.
            \item \textbf{Filter}: Discard UI noise; keep task-relevant static facts.
        \end{itemize}
    \end{minipage}
    \hfill
    \begin{minipage}[t]{0.32\linewidth}
        \textbf{3. VERIFICATION LOGIC RULE}
        \rule{\linewidth}{0.6pt} \\
        \footnotesize
        \begin{itemize}[leftmargin=1.2em, itemsep=0pt, topsep=2pt]
            \item \textbf{Visual Proof}: Compare Image 1 (Before) vs Image 2 (After).
            \item \textbf{Success}: \texttt{click} (state change), \texttt{type} (text visible), \texttt{code} (no crash).
            \item \textbf{Anomaly}: Flag repeated actions or stagnant GUI states ($\ge 2$ steps).
        \end{itemize}
    \end{minipage}

    \vspace{4mm}
    
    \begin{tcolorbox}[
        colback=lightgray, 
        colframe=gray, 
        arc=0mm, 
        boxrule=0.5pt, 
        title=\small\textbf{MANDATORY OUTPUT FORMAT (STRICT COMPLIANCE)}
    ]
        \small\ttfamily
        \textbf{1. State Transition Chain} \\
        \quad [History Narrative: Step 1 $\dots$ Step N-1] \\
        \quad [Current Step Narrative: Step N(Current): `action` $\to$ Outcome] \\
        
        \textbf{2. Cumulative Knowledge Base} \\
        \quad Knowledge \& Constraints: [Fact/Instruction 1], [Fact/Instruction 2] \\
        
        \textbf{3. Latest Action Verification} \\
        \quad Result: [SUCCESS / FAILURE / UNCERTAIN] \\
        \quad Evidence: [Specific visual changes between Image 1 \& 2] \\
        
        \textbf{4. Repetition \& Stagnation Detection} \\
        \quad [Repetition Check] | [Stagnation Check]
    \end{tcolorbox}
    
    \vspace{1mm}
    \centering
    \scriptsize \color{alertRed} 
    \textbf{STRICT CONSTRAINTS:} NO STRATEGIC ADVICE. NO RECOMMENDATIONS. NO INFERENCE BEYOND VISUALS. APPEND-ONLY VERIFIED FACTS.
\end{tcolorbox}
\end{figure*}

\begin{figure*}[t]
\centering
\caption{OBSERVER SYSTEM PROMPT}
\label{fig:obs_prompt}
\begin{tcolorbox}[
    enhanced,
    colback=white,
    colframe=obsGreen,
    arc=0mm,
    boxrule=1.2pt,
    title=\textbf{\textsf{OBSERVER SYSTEM PROMPT}},
    fonttitle=\bfseries\large,
    coltitle=white,
    attach boxed title to top center={yshift=-3mm},
    boxed title style={colback=obsGreen, sharp corners},
    shadow={1.5mm}{-1.5mm}{0mm}{black!10!white} 
]
    \vspace{2mm}
    
    \begin{minipage}[t]{0.32\linewidth}
        \textbf{1. WINDOW \& APP UI STATE}
        \rule{\linewidth}{0.6pt} \\
        \footnotesize
        \begin{itemize}[leftmargin=1.2em, itemsep=0pt, topsep=2pt]
            \item \textbf{Foreground}: Identify the focused window receiving input.
            \item \textbf{Hierarchy}: List all titles, overlapping windows, and taskbar indicators.
            \item \textbf{Modals}: Explicitly describe blocking dialogs, alerts, or file pickers.
        \end{itemize}
    \end{minipage}
    \hfill
    \begin{minipage}[t]{0.32\linewidth}
        \textbf{2. VISIBLE TEXT CONTENT}
        \rule{\linewidth}{0.6pt} \\
        \footnotesize
        \begin{itemize}[leftmargin=1.2em, itemsep=0pt, topsep=2pt]
            \item \textbf{Verbatim}: Transcribe terminal output, editor text, and cell values.
            \item \textbf{UI Labels}: Record menu items, tab names, and sidebar entries.
            \item \textbf{Forms}: Report current values or placeholders in input/search bars.
        \end{itemize}
    \end{minipage}
    \hfill
    \begin{minipage}[t]{0.32\linewidth}
        \textbf{3. INTERACTIVE UI STATE}
        \rule{\linewidth}{0.6pt} \\
        \footnotesize
        \begin{itemize}[leftmargin=1.2em, itemsep=0pt, topsep=2pt]
            \item \textbf{Focus}: Locate the blinking caret, highlighted border, or selection.
            \item \textbf{Progress}: Note spinners, loading bars, or "Processing..." text.
            \item \textbf{Overlays}: Detect tooltips, context menus, or autocomplete lists.
        \end{itemize}
    \end{minipage}

    \vspace{4mm}
    
    \begin{tcolorbox}[
        colback=lightGreen, 
        colframe=obsGreen!30, 
        arc=0mm, 
        boxrule=0.5pt, 
        title=\small\textbf{MANDATORY OBSERVATION OUTPUT FORMAT}
    ]
        \small\ttfamily
        \textbf{1. Application \& Window State} \\
        \quad [Foreground Window | Background Windows | System Dialogs] \\
        
        \textbf{2. Visible Text Content} \\
        \quad [Document/Editor Content | UI Labels | Notifications | Form Values] \\
        
        \textbf{3. Current Interactive State} \\
        \quad [Cursor/Focus | Selection | Scroll Position | Loading State | Overlays]
    \end{tcolorbox}
    
    \vspace{1mm}
    \centering
    \scriptsize \color{alertRed} 
    \textbf{STRICT CONSTRAINTS:} DO NOT SUGGEST NEXT STEPS. DO NOT INFER INTENT. REPORT ONLY WHAT IS DIRECTLY VISIBLE.
\end{tcolorbox}
\end{figure*}

\begin{figure*}[t]
\centering
\caption{PLANNER SYSTEM PROMPT}
\label{fig:plan_prompt}
\begin{tcolorbox}[
    enhanced,
    colback=white,
    colframe=planGray,
    arc=0mm,
    boxrule=1.5pt,
    title=\textbf{\textsf{PLANNER SYSTEM PROMPT}},
    fonttitle=\bfseries\large,
    coltitle=white,
    attach boxed title to top center={yshift=-3mm},
    boxed title style={colback=planGray, sharp corners},
    shadow={2mm}{-2mm}{0mm}{black!15!white} 
]
    \vspace{2mm}
    
    \begin{tcolorbox}[colback=codeBg, colframe=planGray!20, title=\small\textbf{AVAILABLE ACTIONS (Python API)}, sharp corners, boxrule=0.5pt]
        \scriptsize\ttfamily
        \begin{tabular}{lp{11.5cm}}
            \textbf{click}(desc, n, type, keys) & Click described element. n=2 (double) for opening files; n=3 for paragraphs. \\
            \textbf{type}(desc, text, ovr, ent) & Type text. \texttt{overwrite=True} clears field; \texttt{enter=True} submits. \\
            \textbf{drag\_and\_drop}(start, end) & Drag from start description to end description. No coordinates. \\
            \textbf{code}(description) & Execute natural language logic as Python scripts (Calculations/Excel batch). \\
            \textbf{hotkey}(keys) | \textbf{scroll}(desc, n) & Execute combinations (e.g., ['ctrl', 's']) or scroll within elements. \\
            \textbf{done}() | \textbf{fail}() & Signal task completion or trigger replanning on failure/stagnation. \\
        \end{tabular}
    \end{tcolorbox}

    \vspace{1mm}

    \begin{minipage}[t]{0.32\linewidth}
        \textbf{\small 1. ELEMENT DESCRIPTION RULES}
        \rule{\linewidth}{0.6pt} \\
        \tiny
        \begin{itemize}[leftmargin=1em, itemsep=0pt]
            \item \textbf{No Coordinates}: Never output pixels; use visual/functional descriptions.
            \item \textbf{No Special Symbols}: Avoid \texttt{'Sheet1'}, use \textit{Sheet1 tab at bottom}.
            \item \textbf{One Step}: Perform only one verifiable UI action per turn.
        \end{itemize}
    \end{minipage}
    \hfill
    \begin{minipage}[t]{0.32\linewidth}
        \textbf{\small 2. OPERATIONAL LOGIC RULE}
        \rule{\linewidth}{0.6pt} \\
        \tiny
        \begin{itemize}[leftmargin=1em, itemsep=0pt]
            \item \textbf{Verification}: Reopen apps to verify file changes made via code.
            \item \textbf{Recovery}: If Memory reports "REPEATED", call \texttt{agent.fail()} to replan.
            \item \textbf{Success}: Use \texttt{agent.done()} only when the task is fully verified.
        \end{itemize}
    \end{minipage}
    \hfill
    \begin{minipage}[t]{0.32\linewidth}
        \textbf{\small 3. DIRECT CODE CONTROL USAGE}
        \rule{\linewidth}{0.6pt} \\
        \tiny
        \begin{itemize}[leftmargin=1em, itemsep=0pt]
            \item \textbf{Use for}: Batch data entry, precise cell (D2) targeting, or calculations.
            \item \textbf{Prohibited}: No PyAutoGUI, no charts, no visual-only elements.
            \item \textbf{Results}: Raise Exception with \texttt{"RESULT: ..."} to return values.
        \end{itemize}
    \end{minipage}

    \vspace{4mm}
    
    \begin{minipage}[c]{0.55\linewidth}
        \begin{tcolorbox}[colback=logicBlue!5, colframe=logicBlue!30, title=\scriptsize\textbf{MANDATORY OUTPUT FORMAT}, sharp corners]
            \tiny\ttfamily
            \textbf{Thought:} \\
            - Step by Step Progress Assessment (Analyze memory/history) \\
            - Next Action Analysis (Evaluate options \& consequences) \\
            \textbf{Action:} \\
            \texttt{agent.click("Save button in the top toolbar")} \\
            \textbf{Scripts:} (Only for agent.code() actions) \\
            \texttt{[Raw executable Python code]}
        \end{tcolorbox}
    \end{minipage}
    \hfill
    \begin{minipage}[c]{0.42\linewidth}
        \begin{center}
            \color{warnRed} \rule{\linewidth}{1pt} \\
            \scriptsize \textbf{STRICT CONSTRAINT} \\
            \tiny \textbf{NEVER} USE NUMERIC COORDINATES. \\
            \textbf{NEVER} REPEAT FAILED STRATEGIES. \\
            \textbf{OVERWRITE} FILES BY DEFAULT. \\
            \rule{\linewidth}{1pt}
        \end{center}
    \end{minipage}
\end{tcolorbox}
\end{figure*}

%% file: main.bib
@article{wang2024gui,
  title={Gui agents with foundation models: A comprehensive survey},
  author={Wang, Shuai and Liu, Weiwen and Chen, Jingxuan and Zhou, Yuqi and Gan, Weinan and Zeng, Xingshan and Che, Yuhan and Yu, Shuai and Hao, Xinlong and Shao, Kun and others},
  journal={arXiv preprint arXiv:2411.04890},
  year={2024}
}

@article{bai2025qwen3,
  title={Qwen3-vl technical report},
  author={Bai, Shuai and Cai, Yuxuan and Chen, Ruizhe and Chen, Keqin and Chen, Xionghui and Cheng, Zesen and Deng, Lianghao and Ding, Wei and Gao, Chang and Ge, Chunjiang and others},
  journal={arXiv preprint arXiv:2511.21631},
  year={2025}
}

@article{fu2025mano,
  title={Mano Technical Report},
  author={Fu, Tianyu and Su, Anyang and Zhao, Chenxu and Wang, Hanning and Wu, Minghui and Yu, Zhe and Hu, Fei and Shi, Mingjia and Dong, Wei and Wang, Jiayao and others},
  journal={arXiv preprint arXiv:2509.17336},
  year={2025}
}

@inproceedings{nguyen2025gui,
  title={Gui agents: A survey},
  author={Nguyen, Dang and Chen, Jian and Wang, Yu and Wu, Gang and Park, Namyong and Hu, Zhengmian and Lyu, Hanjia and Wu, Junda and Aponte, Ryan and Xia, Yu and others},
  booktitle={Findings of the Association for Computational Linguistics: ACL 2025},
  pages={22522--22538},
  year={2025}
}

@article{song2025coact,
  title={Coact-1: Computer-using agents with coding as actions},
  author={Song, Linxin and Dai, Yutong and Prabhu, Viraj and Zhang, Jieyu and Shi, Taiwei and Li, Li and Li, Junnan and Savarese, Silvio and Chen, Zeyuan and Zhao, Jieyu and others},
  journal={arXiv preprint arXiv:2508.03923},
  year={2025}
}

@misc{yang2026ossymphonyholisticframeworkrobust,
      title={OS-Symphony: A Holistic Framework for Robust and Generalist Computer-Using Agent}, 
      author={Bowen Yang and Kaiming Jin and Zhenyu Wu and Zhaoyang Liu and Qiushi Sun and Zehao Li and JingJing Xie and Zhoumianze Liu and Fangzhi Xu and Kanzhi Cheng and Qingyun Li and Yian Wang and Yu Qiao and Zun Wang and Zichen Ding},
      year={2026},
      eprint={2601.07779},
      archivePrefix={arXiv},
      primaryClass={cs.MA},
      url={https://arxiv.org/abs/2601.07779}, 
}

@inproceedings{cheng2024seeclick,
  title={Seeclick: Harnessing gui grounding for advanced visual gui agents},
  author={Cheng, Kanzhi and Sun, Qiushi and Chu, Yougang and Xu, Fangzhi and YanTao, Li and Zhang, Jianbing and Wu, Zhiyong},
  booktitle={Proceedings of the 62nd Annual Meeting of the Association for Computational Linguistics (Volume 1: Long Papers)},
  pages={9313--9332},
  year={2024}
}

@article{wu2024atlas,
  title={Os-atlas: A foundation action model for generalist gui agents},
  author={Wu, Zhiyong and Wu, Zhenyu and Xu, Fangzhi and Wang, Yian and Sun, Qiushi and Jia, Chengyou and Cheng, Kanzhi and Ding, Zichen and Chen, Liheng and Liang, Paul Pu and others},
  journal={arXiv preprint arXiv:2410.23218},
  year={2024}
}

@article{xie2024osworld,
  title={Osworld: Benchmarking multimodal agents for open-ended tasks in real computer environments},
  author={Xie, Tianbao and Zhang, Danyang and Chen, Jixuan and Li, Xiaochuan and Zhao, Siheng and Cao, Ruisheng and Hua, Toh J and Cheng, Zhoujun and Shin, Dongchan and Lei, Fangyu and others},
  journal={Advances in Neural Information Processing Systems},
  volume={37},
  pages={52040--52094},
  year={2024}
}

@article{jiang2025screencoder,
  title={Screencoder: Advancing visual-to-code generation for front-end automation via modular multimodal agents},
  author={Jiang, Yilei and Zheng, Yaozhi and Wan, Yuxuan and Han, Jiaming and Wang, Qunzhong and Lyu, Michael R and Yue, Xiangyu},
  journal={arXiv preprint arXiv:2507.22827},
  year={2025}
}

@article{agashe2025agent,
  title={Agent s2: A compositional generalist-specialist framework for computer use agents},
  author={Agashe, Saaket and Wong, Kyle and Tu, Vincent and Yang, Jiachen and Li, Ang and Wang, Xin Eric},
  journal={arXiv preprint arXiv:2504.00906},
  year={2025}
}

@article{favreau2026multi,
  title={Do Multi-Agents Dream of Electric Screens? Achieving Perfect Accuracy on AndroidWorld Through Task Decomposition},
  author={Favreau, Pierre-Louis and Lo, Jean-Pierre and Guiguet, Clement and Simon-Meunier, Charles and Dehandschoewercker, Nicolas and Roush, Allen G and Goldfeder, Judah and Shwartz-Ziv, Ravid},
  journal={arXiv preprint arXiv:2602.07787},
  year={2026}
}

@article{chen2024automanual,
  title={Automanual: Constructing instruction manuals by llm agents via interactive environmental learning},
  author={Chen, Minghao and Li, Yihang and Yang, Yanting and Yu, Shiyu and Lin, Binbin and He, Xiaofei},
  journal={Advances in Neural Information Processing Systems},
  volume={37},
  pages={589--631},
  year={2024}
}

@article{dong2025say,
  title={Say One Thing, Do Another? Diagnosing Reasoning-Execution Gaps in VLM-Powered Mobile-Use Agents},
  author={Dong, Lingzhong and Zhou, Ziqi and Yang, Shuaibo and Sheng, Haiyue and Cheng, Pengzhou and Wu, Zongru and Wu, Zheng and Liu, Gongshen and Zhang, Zhuosheng},
  journal={arXiv preprint arXiv:2510.02204},
  year={2025}
}

@article{wanyan2025look,
  title={Look before you leap: A gui-critic-r1 model for pre-operative error diagnosis in gui automation},
  author={Wanyan, Yuyang and Zhang, Xi and Xu, Haiyang and Liu, Haowei and Wang, Junyang and Ye, Jiabo and Kou, Yutong and Yan, Ming and Huang, Fei and Yang, Xiaoshan and others},
  journal={arXiv preprint arXiv:2506.04614},
  year={2025}
}

@article{qin2025ui,
  title={Ui-tars: Pioneering automated gui interaction with native agents},
  author={Qin, Yujia and Ye, Yining and Fang, Junjie and Wang, Haoming and Liang, Shihao and Tian, Shizuo and Zhang, Junda and Li, Jiahao and Li, Yunxin and Huang, Shijue and others},
  journal={arXiv preprint arXiv:2501.12326},
  year={2025}
}

@article{yang2025gta1,
  title={Gta1: Gui test-time scaling agent},
  author={Yang, Yan and Li, Dongxu and Dai, Yutong and Yang, Yuhao and Luo, Ziyang and Zhao, Zirui and Hu, Zhiyuan and Huang, Junzhe and Saha, Amrita and Chen, Zeyuan and others},
  journal={arXiv preprint arXiv:2507.05791},
  year={2025}
}

@article{yin2024survey,
  title={A survey on multimodal large language models},
  author={Yin, Shukang and Fu, Chaoyou and Zhao, Sirui and Li, Ke and Sun, Xing and Xu, Tong and Chen, Enhong},
  journal={National Science Review},
  volume={11},
  number={12},
  pages={nwae403},
  year={2024},
  publisher={Oxford University Press}
}

@article{caffagni2024revolution,
  title={The revolution of multimodal large language models: a survey},
  author={Caffagni, Davide and Cocchi, Federico and Barsellotti, Luca and Moratelli, Nicholas and Sarto, Sara and Baraldi, Lorenzo and Cornia, Marcella and Cucchiara, Rita},
  journal={arXiv preprint arXiv:2402.12451},
  year={2024}
}

@article{agashe2024agent,
  title={Agent s: An open agentic framework that uses computers like a human},
  author={Agashe, Saaket and Han, Jiuzhou and Gan, Shuyu and Yang, Jiachen and Li, Ang and Wang, Xin Eric},
  journal={arXiv preprint arXiv:2410.08164},
  year={2024}
}

@article{zhang2024large,
  title={Large language model-brained gui agents: A survey},
  author={Zhang, Chaoyun and He, Shilin and Qian, Jiaxu and Li, Bowen and Li, Liqun and Qin, Si and Kang, Yu and Ma, Minghua and Liu, Guyue and Lin, Qingwei and others},
  journal={arXiv preprint arXiv:2411.18279},
  year={2024}
}

@article{shaw2023pixels,
  title={From pixels to ui actions: Learning to follow instructions via graphical user interfaces},
  author={Shaw, Peter and Joshi, Mandar and Cohan, James and Berant, Jonathan and Pasupat, Panupong and Hu, Hexiang and Khandelwal, Urvashi and Lee, Kenton and Toutanova, Kristina N},
  journal={Advances in Neural Information Processing Systems},
  volume={36},
  pages={34354--34370},
  year={2023}
}

@article{li2023zero,
  title={A zero-shot language agent for computer control with structured reflection},
  author={Li, Tao and Li, Gang and Deng, Zhiwei and Wang, Bryan and Li, Yang},
  journal={arXiv preprint arXiv:2310.08740},
  year={2023}
}

@inproceedings{hong2024cogagent,
  title={Cogagent: A visual language model for gui agents},
  author={Hong, Wenyi and Wang, Weihan and Lv, Qingsong and Xu, Jiazheng and Yu, Wenmeng and Ji, Junhui and Wang, Yan and Wang, Zihan and Dong, Yuxiao and Ding, Ming and others},
  booktitle={Proceedings of the IEEE/CVF Conference on Computer Vision and Pattern Recognition},
  pages={14281--14290},
  year={2024}
}

@article{xu2024aguvis,
  title={Aguvis: Unified pure vision agents for autonomous gui interaction},
  author={Xu, Yiheng and Wang, Zekun and Wang, Junli and Lu, Dunjie and Xie, Tianbao and Saha, Amrita and Sahoo, Doyen and Yu, Tao and Xiong, Caiming},
  journal={arXiv preprint arXiv:2412.04454},
  year={2024}
}

@article{liu2025infiguiagent,
  title={Infiguiagent: A multimodal generalist gui agent with native reasoning and reflection},
  author={Liu, Yuhang and Li, Pengxiang and Wei, Zishu and Xie, Congkai and Hu, Xueyu and Xu, Xinchen and Zhang, Shengyu and Han, Xiaotian and Yang, Hongxia and Wu, Fei},
  journal={arXiv preprint arXiv:2501.04575},
  year={2025}
}

@article{gonzalez2025unreasonable,
  title={The unreasonable effectiveness of scaling agents for computer use},
  author={Gonzalez-Pumariega, Gonzalo and Tu, Vincent and Lee, Chih-Lun and Yang, Jiachen and Li, Ang and Wang, Xin Eric},
  journal={arXiv preprint arXiv:2510.02250},
  year={2025}
}

@article{cristescu2025ui,
  title={UI-CUBE: Enterprise-Grade Computer Use Agent Benchmarking Beyond Task Accuracy to Operational Reliability},
  author={Cristescu, Horia and Park, Charles and Nguyen, Trong Canh and Talmacel, Sergiu and Ilie, Alexandru-Gabriel and Adam, Stefan},
  journal={arXiv preprint arXiv:2511.17131},
  year={2025}
}

@article{ye2025mobile,
  title={Mobile-agent-v3: Foundamental agents for gui automation},
  author={Ye, Jiabo and Zhang, Xi and Xu, Haiyang and Liu, Haowei and Wang, Junyang and Zhu, Zhaoqing and Zheng, Ziwei and Gao, Feiyu and Cao, Junjie and Lu, Zhengxi and others},
  journal={arXiv preprint arXiv:2508.15144},
  year={2025}
}

@article{huang2025scaletrack,
  title={Scaletrack: Scaling and back-tracking automated gui agents},
  author={Huang, Jing and Zeng, Zhixiong and Han, Wenkang and Zhong, Yufeng and Zheng, Liming and Fu, Shuai and Chen, Jingyuan and Ma, Lin},
  journal={arXiv preprint arXiv:2505.00416},
  year={2025}
}

@article{han2025guirobotron,
  title={GUIRoboTron-Speech: Towards Automated GUI Agents Based on Speech Instructions},
  author={Han, Wenkang and Zeng, Zhixiong and Huang, Jing and Jiang, Shu and Zheng, Liming and Yang, Longrong and Qiu, Haibo and Yao, Chang and Chen, Jingyuan and Ma, Lin},
  journal={arXiv preprint arXiv:2506.11127},
  year={2025}
}

@inproceedings{wu2024autogen,
  title={Autogen: Enabling next-gen LLM applications via multi-agent conversations},
  author={Wu, Qingyun and Bansal, Gagan and Zhang, Jieyu and Wu, Yiran and Li, Beibin and Zhu, Erkang and Jiang, Li and Zhang, Xiaoyun and Zhang, Shaokun and Liu, Jiale and others},
  booktitle={First Conference on Language Modeling},
  year={2024}
}

@article{zhang2025ufo2,
  title={Ufo2: The desktop agentos},
  author={Zhang, Chaoyun and Huang, He and Ni, Chiming and Mu, Jian and Qin, Si and He, Shilin and Wang, Lu and Yang, Fangkai and Zhao, Pu and Du, Chao and others},
  journal={arXiv preprint arXiv:2504.14603},
  year={2025}
}

@article{zhao2025pyvision,
  title={Pyvision: Agentic vision with dynamic tooling},
  author={Zhao, Shitian and Zhang, Haoquan and Lin, Shaoheng and Li, Ming and Wu, Qilong and Zhang, Kaipeng and Wei, Chen},
  journal={arXiv preprint arXiv:2507.07998},
  year={2025}
}

@article{qiu2025alita,
  title={Alita: Generalist agent enabling scalable agentic reasoning with minimal predefinition and maximal self-evolution},
  author={Qiu, Jiahao and Qi, Xuan and Zhang, Tongcheng and Juan, Xinzhe and Guo, Jiacheng and Lu, Yifu and Wang, Yimin and Yao, Zixin and Ren, Qihan and Jiang, Xun and others},
  journal={arXiv preprint arXiv:2505.20286},
  year={2025}
}

@article{wang2025opencua,
  title={Opencua: Open foundations for computer-use agents},
  author={Wang, Xinyuan and Wang, Bowen and Lu, Dunjie and Yang, Junlin and Xie, Tianbao and Wang, Junli and Deng, Jiaqi and Guo, Xiaole and Xu, Yiheng and Wu, Chen Henry and others},
  journal={arXiv preprint arXiv:2508.09123},
  year={2025}
}

@article{xie2025scaling,
  title={Scaling Computer-Use Grounding via User Interface Decomposition and Synthesis},
  author={Xie, Tianbao and Deng, Jiaqi and Li, Xiaochuan and Yang, Junlin and Wu, Haoyuan and Chen, Jixuan and Hu, Wenjing and Wang, Xinyuan and Xu, Yuhui and Wang, Zekun and others},
  journal={arXiv preprint arXiv:2505.13227},
  year={2025}
}

@article{openai2025introducing,
  title={Introducing OpenAI o3 and o4-mini},
  author={OpenAI, Team},
  journal={https://openai. com/index/introducing-o3-and-o4-mini/},
  year={2025}
}

@misc{anthropic20253,
  title={3.7 sonnet and claude code},
  author={Anthropic, Claude},
  year={2025}
}

@article{packer2023memgpt,
  title={MemGPT: Towards LLMs as Operating Systems.},
  author={Packer, Charles and Fang, Vivian and Patil, Shishir\_G and Lin, Kevin and Wooders, Sarah and Gonzalez, Joseph\_E},
  year={2023},
  publisher={ArXiv}
}

@article{li2025screenspot,
  title={Screenspot-pro: Gui grounding for professional high-resolution computer use},
  author={Li, Kaixin and Meng, Ziyang and Lin, Hongzhan and Luo, Ziyang and Tian, Yuchen and Ma, Jing and Huang, Zhiyong and Chua, Tat-Seng},
  journal={arXiv preprint arXiv:2504.07981},
  year={2025}
}

@inproceedings{yao2023react,
  title={React: Synergizing reasoning and acting in language models},
  author={Yao, Shunyu and Zhao, Jeffrey and Yu, Dian and Du, Nan and Shafran, Izhak and Narasimhan, Karthik and Cao, Yuan},
  booktitle={International Conference on Learning Representations (ICLR)},
  year={2023}
}

@article{shinn2023reflexion,
  title={Reflexion: Language agents with verbal reinforcement learning},
  author={Shinn, Noah and Cassano, Federico and Gopinath, Ashwin and Narasimhan, Karthik and Yao, Shunyu},
  journal={Advances in Neural Information Processing Systems},
  volume={36},
  pages={8634--8652},
  year={2023}
}

@article{madaan2023self,
  title={Self-refine: Iterative refinement with self-feedback},
  author={Madaan, Aman and Tandon, Niket and Gupta, Prakhar and Hallinan, Skyler and Gao, Luyu and Wiegreffe, Sarah and Alon, Uri and Dziri, Nouha and Prabhumoye, Shrimai and Yang, Yiming and others},
  journal={Advances in Neural Information Processing Systems},
  volume={36},
  pages={46534--46594},
  year={2023}
}

@article{zhai2024fine,
  title={Fine-tuning large vision-language models as decision-making agents via reinforcement learning},
  author={Zhai, Simon and Bai, Hao and Lin, Zipeng and Pan, Jiayi and Tong, Peter and Zhou, Yifei and Suhr, Alane and Xie, Saining and LeCun, Yann and Ma, Yi and others},
  journal={Advances in neural information processing systems},
  volume={37},
  pages={110935--110971},
  year={2024}
}

@article{bai2025qwen2,
  title={Qwen2. 5-vl technical report},
  author={Bai, Shuai and Chen, Keqin and Liu, Xuejing and Wang, Jialin and Ge, Wenbin and Song, Sibo and Dang, Kai and Wang, Peng and Wang, Shijie and Tang, Jun and others},
  journal={arXiv preprint arXiv:2502.13923},
  year={2025}
}

@article{nguyen2024gui,
  title        = {GUI Agents: A Survey},
  author       = {Nguyen, Dang and Chen, Jian and Wang, Yu and Wu, Gang and Park, Namyong and Hu, Zhengmian and Lyu, Hanjia and Wu, Junda and Aponte, Ryan and Xia, Yu and Li, Xintong and Shi, Jing and Chen, Hongjie and Lai, Viet Dac and Xie, Zhouhang and Kim, Sungchul and Zhang, Ruiyi and Yu, Tong and Tanjim, Mehrab and Ahmed, Nesreen K. and Mathur, Puneet and Yoon, Seunghyun and Yao, Lina and Kveton, Branislav and Kil, Jihyung and Nguyen, Thien Huu and Bui, Trung and Zhou, Tianyi and Rossi, Ryan A. and Dernoncourt, Franck},
  journal      = {arXiv preprint arXiv:2412.13501},
  year         = {2024},
  note         = {Submitted 18 Dec 2024; Revised 26 Sep 2025},
  url          = {https://arxiv.org/abs/2412.13501},
  doi          = {10.48550/arXiv.2412.13501}
}
